\pdfoutput=1
\documentclass{article}

\usepackage[numbers]{natbib}
\usepackage[preprint]{neurips_2024}

\usepackage[utf8]{inputenc} 
\usepackage[T1]{fontenc}    
\usepackage{hyperref}       
\usepackage{url}            
\usepackage{booktabs}       
\usepackage{amsfonts}       
\usepackage{nicefrac}       
\usepackage{microtype}      
\usepackage{xcolor}         

\usepackage{subfigure}
\usepackage{graphicx}
\usepackage{amsmath}
\usepackage{wrapfig}

\title{UniMAP: Universal SMILES-Graph Representation Learning}

\makeatletter
\def\thanks#1{\protected@xdef\@thanks{\@thanks
        \protect\footnotetext{#1}}}
\makeatother

\author{%
    Shikun Feng$^{1}$\thanks{\ $^\ddagger$ Correspondence to \texttt{lanyanyan@air.tsinghua.edu.cn}.} \ , Lixin Yang$^{2}$ \ , Yanwen Huang$^{3}$ \ , Yuyan Ni$^{4}$, \\
  \textbf{Wei-Ying Ma}$^{1}$, \textbf{Yanyan Lan}$^{1}$$^\ddagger$\\
  $^1$Institute for AI Industry Research, Tsinghua University\\
  $^2$Renmin University of China\\
  $^3$Department of Pharmaceutical Science, Peking University\\
  $^4$Academy of Mathematics and Systems Science,
Chinese Academy of Sciences\\
}

\begin{document}

\maketitle

\begin{abstract}
  Molecular representation learning is fundamental for many drug-related applications. Most existing molecular pre-training models are limited in using single molecular modality, either SMILES or graph representation. To effectively leverage both modalities, we argue that it is critical to capture the fine-grained `semantics' between SMILES and graph, because subtle sequence/graph differences may lead to contrary molecular properties. In this paper, we propose a universal SMILE-graph representation learning model, namely UniMAP. Firstly, an embedding layer is employed to obtain the token and node/edge representation in SMILES and graph, respectively. A multi-layer Transformer is then utilized to conduct deep cross-modality fusion. Specially, four kinds of pre-training tasks are designed for UniMAP, including Multi-Level Cross-Modality Masking (CMM), SMILES-Graph Matching (SGM), Fragment-Level Alignment (FLA), and Domain Knowledge Learning (DKL). In this way, both global (i.e.~SGM and DKL) and local (i.e.~CMM and FLA) alignments are integrated to achieve comprehensive cross-modality fusion. We evaluate UniMAP on various downstream tasks, i.e.~molecular property prediction, drug-target affinity prediction, and drug-drug interaction. Experimental results show that UniMAP outperforms current state-of-the-art pre-training methods. We also visualize the learned representations to demonstrate the effectiveness of multi-modality integration.
\end{abstract}

\section{Introduction}
Machine learning has been applied to a wide range of tasks in cheminformatics and bioinformatics, including molecular property prediction~\cite{wu2018moleculenet}, molecular generation~\cite{polykovskiy2020moses}, and virtual screening~\cite{kimber2021dlinvs}. Due to the insufficiency of labeled data and engineered features, representation learning becomes an increasingly important tool in replacement of conventional methods, such as manually designed fingerprints~\cite{morgan1965fingerprint}.

Inspired by the success of pre-training and fine-tuning paradigm, pre-trained models on molecular data have demonstrated promising results. According to data formats, these models can be mainly classified into two categories, SMILES based and graph based methods. SMILES based methods, taking the SMILES~\cite{weininger1988smiles} string as input, have been facilitated by various sequence modeling methods in NLP such as masked language modeling (MLM)~\cite{chithrananda2020chemberta} and auto-encoder~\cite{honda2019smilestrans}. Despite their ease of use, such methods treat molecules as 1D sequences and may fail to capture the geometric structure of a molecule~\cite{rong2020grover}. While graph based methods, taking the molecular graph as input, mainly leverage graph level self-supervised training, such as atom/edge/subgraph masking~\cite{hu2019strategies} and contrastive learning~\cite{wang2022molclr,wang2022imolclr}. These methods directly utilize graph to represent the 2D structure, but may be limited by the over-smoothing~\cite{chen2020smooth} and over-squashing~\cite{topping2022squash} problems of graph neural networks (GNN). 

Given these limitations, a natural idea is to combine SMILES string and molecular graph into one model to enjoy the merits of both approaches. Recently, MOCO~\cite{zhu2022improving} designs a two-stream model to fuse SMILES and graph. Specifically, SMILES and graph are treated as two molecular views, in which a Transformer~\cite{vaswani2017attention} branch and a GNN branch are used to obtain their representations, respectively. Then a consistency loss is conducted on the two molecular-level representations for pre-training. 

\begin{wrapfigure}{r}{0.48\textwidth}
\centering
\includegraphics[scale=0.18]{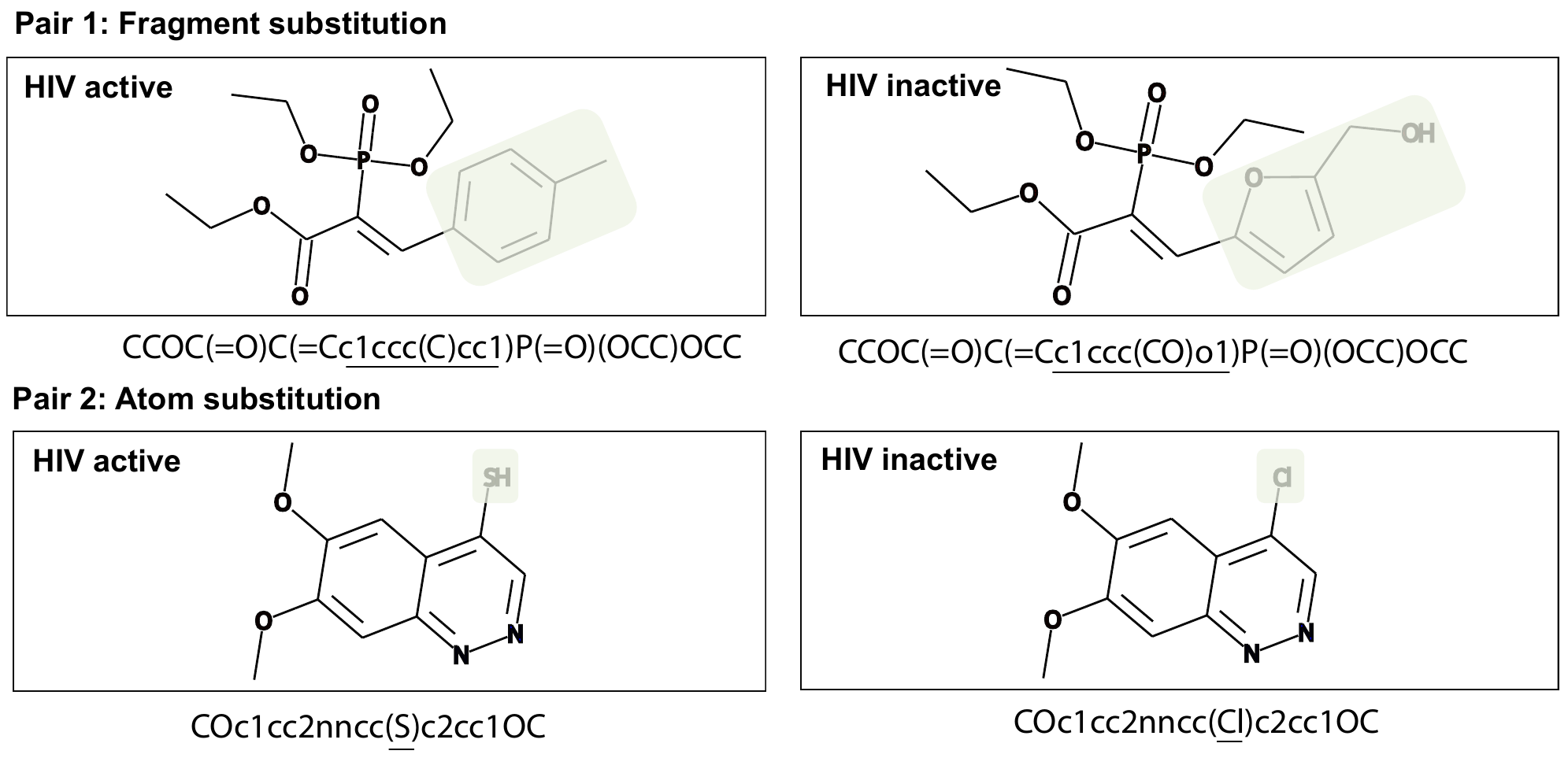}
\caption{Two examples to demonstrate that substitution of a single fragment will lead to opposite properties in HIV dataset from MoleculeNet~\cite{wu2018moleculenet}. Specifically, the green parts in graph and the underlined parts in SMILES indicate the differences between the two molecules. A molecule is labeled as HIV active if it can inhibit HIV replication and HIV inactive otherwise.}
\label{sim_pair}
\end{wrapfigure}

We argue that the above global level alignment cannot well model the correspondence between SMILES and graph. Firstly, SMILES, as a language representation, provides detailed descriptions of a molecular graph, including atoms, bonds, rings, branching, stereochemistry, and isotope. As a result, the global level alignment as in MOCO may fail to recognize the subtle difference between molecules, and cause severe prediction error. For example, the substitution of a single fragment will lead to drug deactivation in virtual screening. We give two example pairs in Figure \ref{sim_pair} for demonstration. In Pair 1, we replace 4-methylbenzyl (SMILES:c1ccc(C)cc1) with 2-hydroxymethylfuran (SMILES:c1ccc(CO)o1), and find that the modified compound no longer inhibits HIV replication. Similarly in Pair 2, if we replace just a single atom, i.e.~thiol (SMILES: S) to chloride (SMILES: Cl), the same result is observed. We have tested previous models on this specific case, including the representative SMILES based method ChemBERTa~\cite{chithrananda2020chemberta} and graph based method GROVER~\cite{rong2020grover}. The similarity results of both ChemBERTa and GROVER are larger than 0.8 for each pair, while UniMAP obtains much smaller similarity results, i.e.~0.55 for Pair 1 and 0.74 for Pair 2. Since the code of MOCO is not released and we cannot reproduce the results in their paper, we do not report the MOCO results on this case. However, given the similarity results from ChemBERTa and GROVER, a single consistency loss will not introduce much difference.

From the above analysis, it is critical to capture the fine-grained `semantics' between SMILES and graph. To this end, we adopt fragments as the basic unit, and propose a multi-modality molecular pre-training model UniMAP, to obtain the UNIversal sMiles-grAPh representation for molecule. Fragments play a very important role in computational pharmaceutical research~\cite{liu2017break,lewell1998recap}. Specifically, molecular fragments are chemically feasible substructures, which to some extent largely determine molecular biological activities and even influence the affinity between a molecule and its target proteins. Therefore, fragments are meaningful substructures to reflect the fine-grained `semantics' between SMILES and graph.

To extract meaningful fragments, we first employ BRICS algorithm~\cite{degen2008art}, to split a molecular graph to different fragments. Then we design a SMILES-graph fragment decomposition algorithm to obtain the corresponding SMILES fragment for each graph fragment. In this way, we could use an embedding layer to obtain both token and fragment representations for a molecule, which act as heterogeneous inputs to facilitate different pre-training tasks. Afterward, a shared Transformer backbone is utilized to conduct deep cross-modality fusion, and output the unified representation for fine-tuning. As for pre-training, besides two molecular-level tasks including SMILES-Graph Matching (SGM) and Domain Knowledge Learning (DKL), we introduce two novel fragment-level tasks to achieve fine-grained cross-modality interactions, i.e.~Multi-Level Cross-Modality Masking (CMM) and Fragment-Level Alignment (FLA). Using both fragments and tokens as masking units, CMM forces the model to acquire both single- and cross-modality information, while FLA uses the fragment embeddings from different modalities as the contrastive learning unit to capture fine-grained cross-modality semantics. In this way, UniMAP learns a better molecular representation by leveraging the fine-grained semantic correlation between SMILES and graph.

We pre-train UniMAP on 10 million molecules from Pubchem~\cite{kim2019pubchem}, and then fine-tune the pre-trained model on various downstream tasks. We achieve state-of-the-art results on molecular property prediction tasks from MoleculeNet and outperform existing methods on drug-target affinity prediction and drug-drug interaction tasks. Our further visualization analysis shows typical fragment alignment patterns and clear embedding clusters, demonstrating the ability of UniMAP to capture both fragment-level and molecular-level `semantics'.


Our main contributions are three folds:
\begin{itemize}
    \vspace{0cm}\item We propose UniMAP, the first single-stream multi-modality molecular representation learning method to leverage both SMILES and graph representation; 
    \vspace{0cm}\item We design both molecular-level (SGM and DKL) and fragment-level (CMM and FLA) pre-training tasks to achieve deep cross-modality fusion in UniMAP; 
    \vspace{0cm}\item We conduct experiments on downstream tasks including molecular property prediction, drug-target affinity prediction, and drug-drug interaction prediction, and demonstrate promising results of UniMAP.
\end{itemize}

\section{Method}
\begin{figure*}[htb] 
\centering
\includegraphics[scale=0.32]{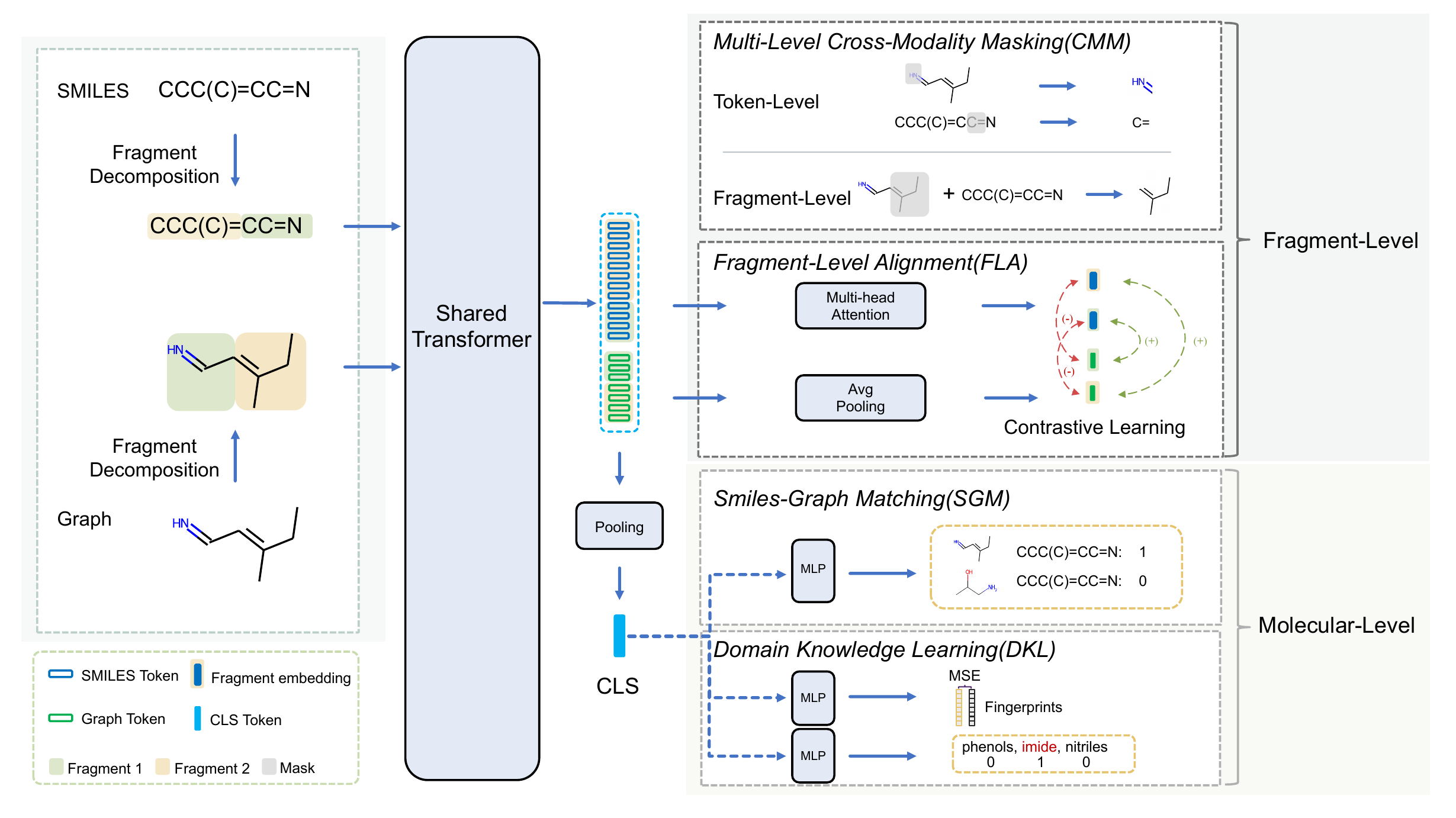}
\caption{Overview of UniMAP. SMILES and Graph are processed by a shared Transformer to get a unified representation, which is supervised by both fragment-level and molecular-level pre-training tasks.}
\label{overall}
\end{figure*}

In this section, we introduce our proposed UniMAP, as shown in Figure~\ref{overall}, including an input embedding layer, a Transformer encoder, and four pre-training tasks.

\subsection{Embedding Layer}
The embedding layer is designed to map the given SMILES and graph of a molecule to embeddings for further computation. For an input SMILES, we adopt a regex-based tokenizer from DeepChem~\cite{Ramsundar-et-al-2019deepchem} to parse it into a series of tokens $S=[t_1, t_2, ..., t_n]$, then we apply an embedding layer to obtain the embeddings of SMILES denoted as $\boldsymbol{s}=[s_1, s_2, ..., s_n]$, where $s_i$ is the embedding of token $i$ with size $D$. While an input graph is denoted as $G=\{V, E\}$, where $V$ is the vertex set with $v_i\in{V}$ denotes the $i$-th atom, and $E$ is the edge set with $e_{ij}\in{E}$ denotes an edge between the $i$-th and the $j$-th atom. We adopt a GCN~\cite{li2020deepergcn} to obtain the graph embeddings $\boldsymbol{g}=[g_1, g_2, ..., g_m]$, where $m=|V|$ is the atom number, and ${g_i}$ is the $D$ dimensional vector for the $i$-th atom.

\subsection{Transformer Encoder}

We adopt a Transformer based encoder denoted as $\boldsymbol{\theta}$ to handle both SMILES and graph embeddings. The typical Transformer takes a sequence of embeddings as input. We add a learnable position embeddings denoted as $p_{\boldsymbol{s}}$ to embeddings of SMILES $\boldsymbol{s}$ to maintain positional information. As for the molecular graph in which atoms are not organized in sequence, we simply concatenate graph embeddings $\boldsymbol{g}$ after the SMILES tokens embeddings to obtain the input of the Transformer,  denoted as $\boldsymbol{z}$.
\begin{small}
\begin{equation}
\boldsymbol{z} = [\boldsymbol{s} + p_{\boldsymbol{s}},\boldsymbol{g}]. \end{equation}
\end{small}

The Transformer encoder is composed of stack of identical blocks, where each block consists of two parts: a multi-head self-attention (MSA) module and a feed-forward neural network (FFN). Specifically, the self-attention mechanism in MSA allows cross-modality attention and conducts multi-modality interaction, including both inter- and intra-modality fusions.


To this end, the embeddings $\boldsymbol{x}= \boldsymbol{\theta
}(\boldsymbol{z})$ is obtained for each molecule, where $\boldsymbol{x}=(x_1, x_2, ..., x_n, x_{n+1},... ,x_{n+m}), x_i \in {\mathbb{R}}^{D}$, in which the first $n$ elements and the last $m$ elements of $\boldsymbol{x}$ stand for the corresponding SMILES and graph representations.

Afterward, we apply a typical average pooling operation~\cite{huang2021whiteningbert} on $\boldsymbol{x}$ to get the final molecular representation, denoted as $x_{cls}$. 
\begin{small}
\begin{equation}
x_{cls}=\frac{1}{m+n} \sum_{i=1}^{m+n}{x_i}.   
\end{equation}
\end{small}

Based on $x_{cls}$, different losses could be designed to facilitate the pre-training process. Considering the characteristics of SMILES and graph relation, we design both local (i.e.~fragment-level) and global (i.e.~molecular-level) losses.



\subsection{SMILES-Graph Fragment Decomposition}
Our SMILES-graph fragment decomposition algorithm first applies a mature method to decompose graph into different fragments, then assigns each SMILES character to the corresponding fragment, based on the SMILES definition.

Specifically, we use BRICS~\cite{degen2008art} algorithm to split the molecular graph into different fragments, i.e.~$K_j$ fragments obtained for the $j$-th molecule, denoted as $K$ for convenience. The labels of graph nodes could be represented as an vector $\boldsymbol{l}_g = [l_1^{(g)}, l_2^{(g)}, ..., l_m^{(g)}]$ to demonstrate which fragment each atom belongs to, where $l_i^{(g)}$ stands for the fragment ID of the $i$-th atom with value from $0$ to $K-1$.

Since SMILES contains not only atoms but also special symbols representing chemical meanings, we define 4 rules as follows to label the special symbols, based on their meanings in SMILES grammar~\cite{weininger1988smiles}. Each character of $\{., -, =, \#, \$, :, /, \backslash \}$ is assigned the same label with its left nearest atom, because they denote the type of the corresponding bond started from the left nearest atom. The number after the character `C' is assigned the same label as the character `C', because the number usually indicates the beginning or the ending of a ring structure. The special sub-string `@@' and `@' are assigned the same fragment label with the carbon they decorate. Parentheses and brackets are assigned the same label as the nearest surrounding atom, because the contents inside them are usually in the same fragment. After label assignment, the labels of SMILES tokens are denoted as $\boldsymbol{l}_s = [l_1^{(s)}, l_2^{(s)}, ..., l_n^{(s)}]$, where  $l_i^{(s)}$ stands for the fragment ID of the $i$-th token with value from $0$ to $K-1$.

\subsection{Multi-Level Cross-Modality Masking}

We design two kinds of cross-modality mask language modeling methods in this paper, i.e.~token-level masking and fragment-level cross-modality masking. For fragment-level masking, the goal is similar to the previous conditional masking strategy~\cite{chen2020uniter}, i.e~using the surrounding contexts and the full information from the other modality to recover the corresponding masked SMILES or graph fragments. However, for tokens, we define the goal as only using the surrounding context of the single modality itself to predict the masked SMILES tokens or graph atoms, to emphasize the intra-modality pattern. In this way, both inter- and intra-modality patterns could be considered. This is reasonable because tokens are relatively smaller than fragments, and their context information in the same modality could be sufficient for MLM. Comparison with these masking strategies is conducted in the ablation study.


Firstly, we introduce the token-level masking. The input tokens of SMILES $S$ and atoms of graph $G$ are masked independently with the same ratio $r_t$. Specifically, for the SMILES, we randomly select $|S|*r_t$ tokens from SMILES $S$ to mask and predict the type of masked tokens. As for the graph, we randomly select $|V|*r_t$ atoms, set the initial feature of both selected atoms and their adjacent edges to the pre-defined masked ones, and then utilize GNN on the masked graph to obtain node embeddings. We follow the practice in GROVER to predict the masked atom's contextual information, and the total loss is defined as follows.


\begin{small}
\begin{equation}
    \mathcal{L}_{t}  = - \left( \sum_{\forall S} (\log P_{\boldsymbol\theta}(\boldsymbol{s_m}|S_{\backslash \boldsymbol{m}}) ) + \sum_{\forall G} ( \log P_{\boldsymbol\theta}(\boldsymbol{g_m}| G_{\backslash \boldsymbol{m}} )) \right ),
\end{equation}
\end{small}

where $\boldsymbol{s_m}$ and $\boldsymbol{g_m}$ stand for masked SMILES tokens, atoms and edges, respectively. $S_{\backslash \boldsymbol{m}}$ and $G_{\backslash \boldsymbol{m}}$ stand for the surrounding contexts in the SMILES $S$ and graph $G$.


For a given SMILES-graph pair $(S, G)$, we introduce the fragment-level cross-modality masking (f-CMM). In this approach, we randomly select $r_f * K$ fragments and then randomly choose one modality with a probability of 0.5 to mask the selected fragments. The data processing and predicting target are the same with token-level masking except for taking fragments as masking unit. Thus the loss of f-CMM can be written as:
\begin{small}
\begin{equation}
    \mathcal{L}_{f} = -\sum_{\forall (S,G)} ( \log P_{\boldsymbol\theta}(\boldsymbol{s_m}|S_{\backslash \boldsymbol{m}}, G) + \log P_{\boldsymbol\theta}(\boldsymbol{g_m}|S, G_{\backslash \boldsymbol{m}}) ),
\end{equation}
\end{small}
where $\boldsymbol{s_m}$ and $\boldsymbol{g_m}$ stand for masked SMILES and graph fragments, respectively. $S_{\backslash \boldsymbol{m}}$ and $G_{\backslash \boldsymbol{m}}$ stand for the surrounding contexts in the SMILES $S$ and graph $G$.

The overall multi-level cross-modality masking loss, denoted as $\mathcal{L}_{CMM}$, is the sum of $\mathcal{L}_{t}$ and $\mathcal{L}_{f}$:
\begin{small}
\begin{equation}
    \mathcal{L}_{CMM} = \mathcal{L}_{t} + \mathcal{L}_{f} .
\end{equation}
\end{small}






\vspace{-0.2cm}
\subsection{Fragment-Level Alignment}

Fragment-level alignment (FLA) is a fine-grained cross-modality alignment strategy. The basic idea is to obtain corresponding fragment representations in a SMILES and graph pair, and then align them by a contrastive loss. 

For each fragment $i=0,\cdots, K-1$, the SMILES representation $f_i^{(s)}$ is obtained by a multi-head attention layer with average pooling to aggregate the associated token embeddings, and the graph representation $f_i^{(g)}$ is directly obtained by aggregating the associated atom embeddings with average pooling. Clearly, $f_i^{(s)}$ and $f_i^{(g)}$ form a positive pair because they stand for the same fragment in different modalities. While the negative pairs are constructed as follows. For each fragment $f_i^{(s)}$, the negative graph fragment set $N_s$ includes both other fragments from the same graph and all fragments from other graphs. Similarly, we can obtain the negative SMILES fragment set $N_g$ for each fragment $f_i^{(g)}$.

Afterward, we define the fragment-level contrastive learning loss as a combination of SMILES specific and graph specific contrastive losses, i.e.~$\mathcal{L}_{FLA} = \mathcal{L}_{s} + 
 \mathcal{L}_{g}$, with

\begin{small}
\begin{equation}
\mathcal{L}_{s} = - \sum_{i} ( \log  \frac{e^{cos(f_i^{(s)}, f_i^{(g)}) / \tau}}{e^{cos(f_i^{(s)}, f_i^{(g)}) / \tau} + \sum_{j \in N_s} e^{cos(f_i^{(s)}, f_j^{(g)}) / \tau}} ),
\end{equation}

\begin{equation}
\mathcal{L}_{g} = - \sum_{i} ( \log \frac{e^{cos(f_i^{(g)}, f_i^{(s)}) / \tau}}{e^{cos(f_i^{(g)}, f_i^{(s)}) / \tau} + \sum_{j \in N_g} e^{cos(f_i^{(g)}, f_j^{(s)}) / \tau}} ),
\end{equation}
\end{small}
where $cos(,)$ represents the cosine similarity function and $\tau$ is the temperature hyper-parameter.

\subsection{SMILES-Graph Matching}
SMILES-Graph Matching is designed to reflect the molecular-level cross-modality alignment. Specifically, for each positive SMILES-graph pair $(S,G)$, we construct an negative pair $(S,G')$ by randomly replacing
the graph $G$ to another molecular graph $G'$ in the training set, and the corresponding molecular representations $x_{cls}$ and $x_{cls}'$ are then taken as input to an MLP head $\boldsymbol{\theta}_{sg}$ to predict the corresponding binary label 1 and 0, with loss as follows.



\begin{small}
\begin{equation}
\mathcal{L}_{SGM} = -\left(\sum_{\forall (S,G)}\log P_{\boldsymbol{\theta}_{sg}}(x_{cls}) + \sum_{\forall (S,G')}\log(1-P_{\boldsymbol{\theta}_{sg}}(x_{cls}')\right).
\end{equation}
\end{small}

\subsection{Domain Knowledge Learning}
We also design two pre-training losses to learn important chemical domain knowledge. Specifically, an MLP layer is added to predict the extended-connectivity fingerprints~\cite{rogers2010extended}, which contains information about molecular topological structure and activity. The Mean Squared Error (MSE) loss is utilized for optimization. Furthermore, we add a classification layer to predict functional groups in a molecule, as in GROVER. The domain knowledge learning loss is defined as follows:

\begin{small}
\begin{equation}
\mathcal{L}_{DKL} = \sum_{\forall (S,G)}||\boldsymbol{\theta}_{fp}(x_{cls}) - y_{fp}||_{2}^{2} - \sum_{\forall (S,G)}\log P_{\boldsymbol{\theta}_{fg}}(y_{fg}|x_{cls}),
\end{equation}
\end{small}

where $\boldsymbol{\theta}_{fp}$ and $\boldsymbol{\theta}_{fg}$ represent the MLP head, while $y_{fp}$ and $y_{fg}$ denote the labels of fingerprints and function groups respectively.



Finally, we set the weight of each loss to 1 due to stable optimization observations of their combination. Thus, the overall pre-training loss $L$ is written as:
\begin{small}
\begin{equation}
\mathcal{L} = \mathcal{L}_{CMM} + \mathcal{L}_{FLA} + \mathcal{L}_{SGM} + \mathcal{L}_{DKL}
\end{equation}
\end{small},
where $\mathcal{L}_{DKL}$ denotes the domain knowledge learning loss.

\section{Experiments} \label{experiments}
This section introduces our experimental evaluation of UniMAP on three different kinds of downstream tasks, including molecular property prediction, drug-target affinity prediction, and drug-drug interaction prediction(DDI). The results for DDI are detailed in appendix~\ref{ddi section}

For pre-training, we select about 10 million molecules from PubChem, which is a widely used and publicly accessible database in chemistry, it contains about 114 million compounds which mostly are small molecules. In the data preprocessing phase, SMILES are canonicalized and converted to molecular graph by RDKit~\cite{landrum2013rdkit}. In the training phase, we adopt the RoBERTa model as the multi-modality encoder $\boldsymbol{\theta}$ and utilize Adam for optimization. Specifically, we utilize the Roberta base model as the shared transformer which contains 12 layers and the hidden size of embedding
is set to 768. For optimization, the initial learning rate is set to 0.00005, and is then decreased by a linear decay scheduler. The mask ratio
of token-level masking and fragment-level cross-modality
masking is set as 0.2 and 0.6 separately, and the temperature
$\tau$ of contrastive learning in fragment-level alignment is set
as 0.05. Since we have multiple pre-training losses, the model is trained in a multi-task learning manner, and the weight of each per-training task is equal to 1. We train
our model for 40 epochs on an 8×V100 GPUs machine. The
batch size per GPU is set to 64. In total, it takes about 10 days to
finish the training process, and half-precision training is utilized to save GPU memory.

\begin{table*}[t]
    \caption{Results in terms of mean and std for the 8 classification tasks from MoleculeNet benchmark, we report ROC-AUC(\%) performance. The best and second best results are marked \textbf{bold} and \underline{underlined}, respectively. `-' denotes the case that the original results are not provided.}
    \centering
    \scalebox{0.73}{
    \begin{tabular}{llllllllll}
    \hline
        \textbf{Method} & \textbf{BBBP} & \textbf{Sider} & \textbf{ClinTox} & \textbf{Bace} & \textbf{Tox21} & \textbf{Toxcast} & \textbf{HIV} & \textbf{MUV} & \textbf{Average} \\ \hline
        ChemBERTa &64.3(-)&-&73.3(-)&-&72.8(-)&-&62.2(-)&-&68.1 \\
        SMILES Transformer &70.4(-)&-&-&70.1(-)&-&-&72.9(-)&-&71.1 \\ \hline \hline
        KCL & 69.9(0.6) & 63.6(0.8) & 58.8(1.9) & 80.2(1.8) & 70.6(0.5) & 63.8(0.1) & 75.7(0.6) & 70.4(0.9) & 69.1 \\ 
        GROVER(base) & 71.9(0.1) & 65.6(0.9) & \underline{81.7(2.5)} & 83.3(0.1) & 73.6(0.3) & \underline{66.1(0.1)} & 75.0(0.2) & 76.7(0.2) & 74.2 \\ 
        GROVER(large) & 71.8(0.5) & \textbf{66.8(0.1)} & 73.9(5.6) & \underline{83.5(0.1)} & 73.4(0.1) & 65.9(0.2) & 73.1(1.7) & 72.1(5.5) & 72.6 \\ 
        AttrMasking & 64.3(2.8) & 61.0(0.7) & 71.8(4.1) & 79.3(1.6) & 76.7(0.4) & 64.2(0.5) & 77.2(1.1) & 74.7(1.4) & 71.2 \\ 
        ContextPred & 68.0(2.0) & 60.9(0.6) & 65.9(3.8) & 79.6(1.2) & 75.7(0.7) & 63.9(0.6) & 77.3(1.0) & 75.8(1.7) & 70.9 \\ 
        GraphLoG & 72.5(0.8) & 61.2(1.1) & 76.7(3.3) & 83.5(1.2) & 75.7(0.5) & 63.5(0.7) & 77.8(0.8) & 76.0(1.1) & 73.4\\ 
        GraphMAE  & 72.0(0.6) & 60.3(1.1) & 82.3(1.2) & 83.1(0.9) & 75.5(0.6) & 64.1(0.3) & 77.2(1.0) & 76.3(2.4) & 73.8 \\ 
        MGSSL  & 70.5(1.1) & 60.5(0.7) & 79.7(2.2) & 79.7(0.8) & 76.4(0.4) & 63.8(0.3) &  \underline{79.5(1.1)} & 78.1(1.8) & 73.5 \\ 
        Mole-BERT  & 72.3(0.7) & 62.2(0.8) & 80.1(3.6) & 81.4(1.0) & \underline{77.1(0.4)} & 64.5(0.4) & 78.6(0.7) & 78.3(1.2) & 74.3\\
        \hline \hline
        GraphMVP & 72.4(1.6) & 63.9(1.2) & 79.1(2.8) & 81.2(0.9) & 75.9(0.5) & 63.1(0.4) & 77.0(1.2) & 77.7(0.6) & 73.8 \\ 
        3D InfoMax & 69.1(1.1) & 53.4(3.3) & 59.4(3.2) & 79.4(1.9) & 74.5(0.7) & 64.4(0.9) & - & 76.1(1.3) & 68.0 \\
        
        \hline \hline
        MOCO  & 71.6(1.0) & 61.2(0.6) & 81.6(3.7) & 82.6(0.3) & 76.7(0.4) & 64.9(0.8) & 78.3(0.4) & \underline{78.5(1.4)} & \underline{74.4}\\ \hline
        \textbf{UniMAP} & \textbf{75.6(1.8)} & \underline{66.6(0.8)} & \textbf{98.3(1.7)} & \textbf{83.6(1.0)} & \textbf{77.2(0.9)} & 
 \textbf{67.0(0.4)} & \textbf{80.1(0.7)} & \textbf{78.9(1.4)} & \textbf{78.4} \\ \hline
    \end{tabular}
    }
    \label{molnetcls}
\end{table*}
\vspace{-0.2cm}
\subsection{Molecular Property Prediction}

\textbf{Data.} We experiment on MoleculeNet~\cite{wu2018moleculenet}, a widely used benchmark, to test the molecular property prediction ability, including three biophysics properties(i.e.~Bace, HIV, and MUV) and five physiology properties (i.e.~BBBP, Sider, ClinTox, Tox21, and Toxcast). Following previous practices in MoleculeNet, each dataset is split to train, validation, and test set with a ratio of 8:1:1, based on the scaffold splitter of DeepChem. Since the targets are classification tasks, we use ROC-AUC(\%) as the evaluation metric. Specially for UniMAP, we run 3 times with different random seeds for each dataset and report the mean and standard deviation values.




\textbf{Baseline Methods.} We compare UniMAP with various pre-training baselines, including SMILES based methods ChemBERTa~\cite{chithrananda2020chemberta} and SMILES Transformer~\cite{honda2019smilestrans}, graph based methods KCL~\cite{fang2022kcl}, GROVER~\cite{rong2020grover}, AttrMasking~\cite{hu2019strategies}, ContextPred~\cite{hu2019strategies}, GraphLoG~\cite{xu2021self}, GraphMAE~\cite{hou2022graphmae}, MGSSL~\cite{zhang2021mgssl}, and Mole-BERT~\cite{anonymous2023molebert}, 3D based methods 3D InfoMax~\cite{stark20223d} and GraphMVP~\cite{liu2022graphmvp}, and hybrid method MOCO~\cite{zhu2022improving}. Most results are from the original paper except KCL and GROVER, since they use different data split methods. To guarantee a fair comparison, we employ their pre-trained models and conduct fine-tuning on the same data with other baselines for evaluation.

\textbf{Fine-Tuning UniMAP.}
Based on the pre-trained representation $x_{cls}$, a two-layer MLP is adopted to obtain the final classification label, supervised by a corresponding classification loss, i.e.~cross entropy. For each task, we try 20 different hyper-parameter combinations, including learning rate, batch size, and weight-decay, via grid search on the validation set and report the best results.




\textbf{Results.} As shown in Table~\ref{molnetcls}, UniMAP achieves the best results on 7 out of 8 datasets, while on Sider our result of 66.6\% is the second best, which is slightly worse than GROVER's 66.8\%. Specifically from the averaged results, we can see that UniMAP (i.e.~78.4\%) significantly outperforms the second-best method MOCO (i.e.~74.4\%), which is also a hybrid method that combines four different molecular data forms. These results confirm the strength of UniMAP to perform consistently better on various molecular property prediction datasets from MoleculeNet, by using both SMILES and graph representation with fine-grained alignment.




\subsection{Drug-Target Affinity Prediction}

\begin{wraptable}{r}{0.5\textwidth}
\caption{Experiment results on DTA benchmarks. We report MSE and Concordance Index (CI) scores, the best and second best results are marked \textbf{bold} and \underline{underlined}, respectively.}
\centering
\scalebox{0.9}{
\begin{tabular}{lllll}
\hline
\textbf{Task}   & \multicolumn{2}{l}{\textbf{Davis}}          & \multicolumn{2}{l}{\textbf{KIBA}}           \\
\textbf{Method} & \textbf{MSE↓} & \textbf{CI↑} & \textbf{MSE↓} & \textbf{CI↑} \\ \hline
KronRLS   & 0.329  & 0.847  & 0.852 & 0.688 \\
GraphDTA  & 0.263  & 0.864  & 0.183 & 0.862 \\
DeepDTA   & 0.262  & 0.870   & 0.196 & 0.864 \\
SMT-DTA   & \textbf{0.219} & \textbf{0.890} & \underline{0.154} & \textbf{0.894} \\
GraphMVP  & 0.274  & -      & 0.175 & -     \\
Mole-BERT & 0.266  & -      & 0.157 & -     \\
3D PGT & 0.275 & - & 0.173 & - \\
SimSGT & 0.251 & - & 0.153 & - \\
\hline
\textbf{UniMAP}  & \underline{0.246} & \underline{0.888} & \textbf{0.144} & \underline{0.891} \\ \hline
\end{tabular}} %
\label{dti}
\end{wraptable}
\textbf{Data.}
Drug-Target Affinity (DTA) prediction is an important task in drug discovery, which is designed to predict the binding affinity between a drug and a target protein. We evaluate our method on two widely used DTA benchmarks, DAVIS~\cite{davis2011comprehensive} and KIBA~\cite{tang2014making}. DAVIS records the dissociation constant($K_d$) value of kinase protein and its inhibitors, while KIBA contains the $K_i$, $K_d$, and $IC_{50}$ values to describe the bioactivity of kinase inhibitors. We follow GraphMVP to split the datasets and calculate the MSE and Concordance Index (CI) scores as evaluation metrics.

\textbf{Baseline Models.}
We compare our method with three different types of baselines. KronRLS~\cite{panico1993iupac}, GraphDTA~\cite{nguyen2021graphdta} and DeepDTA~\cite{ozturk2018deepdta} are supervised methods for DTA; SMT-DTA~\cite{pei2022smt} is a semi-supervised method, which trains the model with an additional unsupervised masked language modeling task with large-scale unlabeled molecules and protein data; GraphMVP, 3D PGT~\cite{wang2023automated},  SimSGT~\cite{liu2024rethinking} and Mole-BERT are molecular pre-training methods.

\textbf{Fine-Tuning UniMAP.}
Similarly to GraphMVP, we model the target protein with a convolution neural network to get the protein embedding, concatenate it with the molecular embedding $x_{cls}$, and then use an MLP to obtain the predicted binding affinity value.

\textbf{Results.}
The results are demonstrated in Table~\ref{dti}. We report MSE and Concordance Index(CI) scores as evaluation metrics, GraphMVP and Mole-BERT didn't provide CI scores in their original results. From the results, we can see that UniMAP outperforms all supervised DTA methods and other molecular pre-training methods, and performs slightly worse than SMT-DTA. It may be because SMT-DTA utilized large-scale unlabeled protein data and a more complicated Transformer to learn the protein embedding. Still, UniMAP performs better in terms of MSE in KIBA, indicating the ability of UniMAP to learn a meaningful representation for DTA.

\vspace{-0.2cm}
\section{Discussions}
In this section, we provide an ablation study of losses and visualization results, to facilitate further understanding of UniMAP.

\subsection{Ablation Study on Different Losses}

\begin{table*}[h]
    \caption{Comparison with different loss combinations.}
    \centering
    \scalebox{0.68}{
    \begin{tabular}{llllllllll}
    \hline
        \textbf{Method} & \textbf{BBBP} & \textbf{Sider} & \textbf{ClinTox} & \textbf{Bace} & \textbf{Tox21} & \textbf{Toxcast} & \textbf{HIV} & \textbf{MUV} & \textbf{Average} \\ \hline
        SGM + CMM + FLA + DKL  & \textbf{76.1(0.4)} & \textbf{65.8(0.1)} & \underline{98.6(0.2)} & \underline{80.9(0.3)} & \textbf{76.3(0.6)} & \underline{60.1(0.4)} & \underline{77.1(0.4)} & \underline{72.6(0.3)} & \textbf{75.9} \\ 
        SGM + CMM +FLA & \underline{75.0(0.4)} & 63.6(0.4) & \textbf{98.7(0.2)} & \textbf{81.2(0.8)} & 74.9(0.3) & \textbf{60.8(0.8)} & \textbf{77.6(1.5)} & 71.8(1.4) & \underline{75.5} \\ 
        SGM + CMM & 71.1(0.1) & \underline{63.8(0.3)} & 98.1(0.4) & 80.1(0.8) & 75.4(0.2) & 59.4(0.6) & 76.5(0.9) & \textbf{72.8(1.1)} & 74.6 \\ 
        SGM +  Conditional Masking & 71.1(0.5) & 60.7(0.6) & 96.4(0.8) & 79.3(0.3) & 74.6(0.1) & 58.1(1.8) & 72.2(0.3) & 67.7(0.7) & 72.5 \\ 
        SGM +  Single-Modality Masking & 71.0(0.3) & 61.7(0.4) & 95.8(0.7) & 79.2(1.4) & \underline{75.6(0.2)} & 57.4(0.4) & 75.7(0.1) & 69.3(0.7) & 73.2 \\ \hline
    \end{tabular} }
\label{abi}
\end{table*}

Our ablation study contains two parts, comparing different losses, and different masking strategies. We test on 8 typical classification tasks, and the results are shown in Table~\ref{abi}. Each setting pre-trains the model using only 1 million molecules for the consideration of saving time and computational resources.


Firstly, we delete DKL and FLA one by one. From the results,
we can see that DKL proves particularly beneficial for tasks such as Sider and Tox21. This advantage can be attributed to the function group prediction task within DKL, which directly relates to the molecular side effects and toxicity properties. Certain functional groups, when present, can induce adverse reactions or toxicity in the human body. For example, the presence of aldehyde groups has been linked to hepatocyte side effects~\cite{lopachin2014molecular}. Fragment-Level Alignment(FLA) is particularly advantageous for tasks like BBBP and HIV. The alignment of fragments across different modalities enables the model to explicitly capture the semantics of fragments, thus enhancing sensitivity to fragment replacement which may influence molecular properties. As depicted in Figure 1, substituting a fragment can lead to HIV inhibition. Similarly, in BBBP, pharmacologists commonly examine fragments within a drug to either promote or inhibit blood-brain barrier permeability~\cite{xiong2021strategies}.


Along with the SGM task which is commonly adopted in existing multi-modal methods, we compare CMM in UniMAP with the other two token-level mask strategies, i.e.~token-level conditional masking and single-modality masking. Specifically, token-level conditional masking means that we only mask one-modality tokens, i.e.~SMILES tokens or graph atoms, to conduct the prediction based on two modality data; while single-modality masking means that the model predicts the masked tokens from only the surrounding tokens of the same modality. The results show that our CMM performs the best on most tasks, except on Tox21 (i.e.~slightly worse than conditional masking). The overall performance superiority of CMM over the other two masking strategies can be attributed to our proposed more challenging fragment-level conditional masking and the consideration of both intra-modal and inter-modal patterns. Therefore, we can see that an in-depth multi-modality dependency strategy will produce a better molecular representation.



\subsection{Embedding Clustering Results}
We conduct visualization analyses on the output embeddings from both molecular and fragment levels. 

\begin{figure}[t]
\centering
\begin{minipage}{0.4\textwidth}
    \centering
    \includegraphics[scale=0.2]{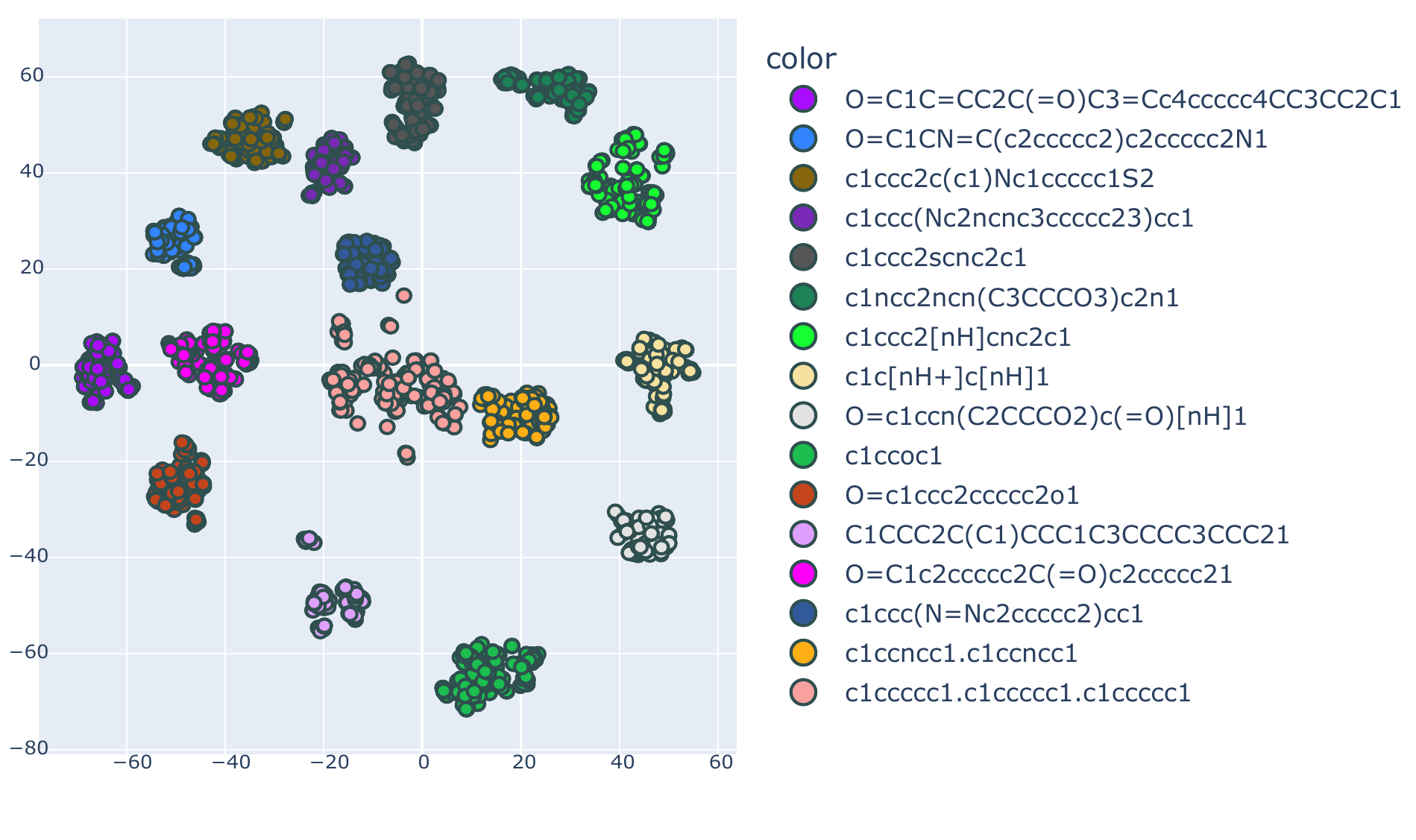}
    \vspace{-0.2cm}
    \caption{Visualization of molecular embeddings with colors indicating different scaffolds.}
    \label{mol_cluster}
\end{minipage}
\hfill
\begin{minipage}{0.49\textwidth}
    \centering
    \includegraphics[scale=0.2]{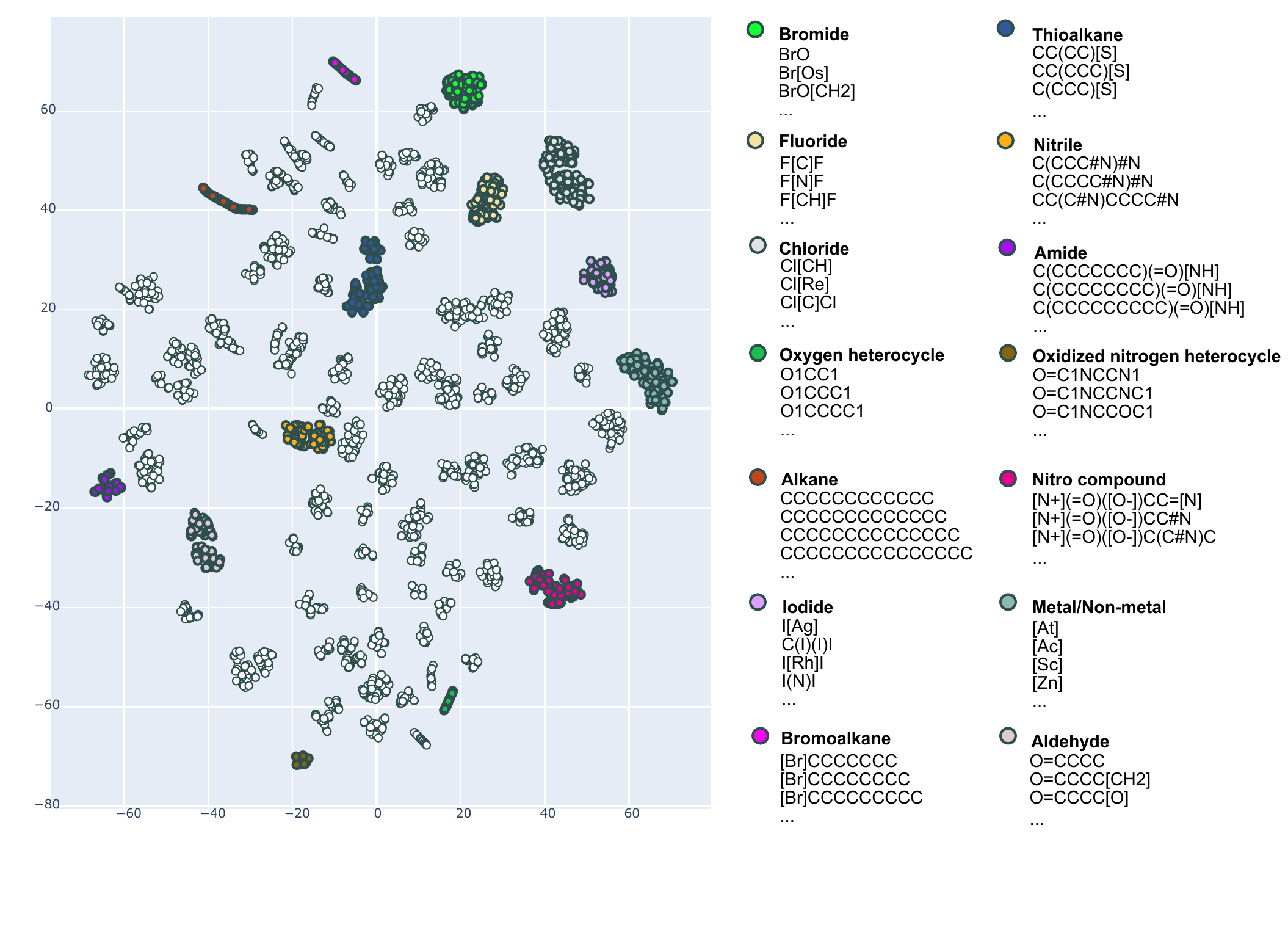}
    \vspace{-1.2cm}
    \caption{Visualization of fragment embeddings. Several clusters are labeled with colors indicating different categories, of which chemical names following several representative fragments are listed on the right.}
    \label{fragment_cluster}
\end{minipage}
\end{figure}

Firstly, we randomly select about two thousand molecules belonging to pre-defined 16 types of scaffold (denoted as different colors), then visualize their UniMAP representations through t-SNE~\cite{van2008visualizing}. The results are shown in Figure \ref{mol_cluster}. We can see that the learned representations demonstrate a clear scaffold clustering property, i.e.~molecules with the same scaffold are mapped to close positions in the embedding space. 

In addition, we visualize the embeddings of four thousand molecular fragments, as shown in Figure \ref{fragment_cluster}. We find that some clustered fragments correspond to certain chemical categories~\cite{clayden2012organic}. For example, four adjacent clusters colored in light yellow, light grey, bright green, and magenta stand for Fluoride, Chloride, Bromide, and Iodide respectively, which contain different atoms of halogens like F, Cl, Br, and I. Three clusters colored red, blue, and magenta imply Alkane, Thioalkane, and Bromoalkane respectively. We also find three nitrogenous fragment clusters including Nitrile, Amide, and Nitro compound, and two heterocycle clusters including Oxygen heterocycle and Oxidized nitrogen heterocycle. 


From the above visualization results, we can conclude that UniMAP provides meaningful molecular-level and fragment-level representations.




\vspace{-0.3cm}

\section{Conclusion}
In this paper, we propose the single-stream multi-modality model to fusion both SMILES and graph in a fine-grained manner for molecular representation learning, namely UniMAP. Firstly, a SMILES-graph decomposition algorithm is designed to obtain corresponding SMILES and graph fragments, which together with tokens function as heterogenous input in pre-training tasks. Secondly, a shared Transformer is utilized to conduct deep cross-modality fusion. Besides molecular-level tasks such as SMILES-Graph Matching and Domain Knowledge Learning, we introduce two novel fragment-level pre-training tasks, i.e. Multi-Level Cross-Modality Masking and Fragment-Level Alignment. Finally, we conduct extensive experiments on three different kinds of downstream tasks, including molecular property prediction, drug-target affinity prediction, and drug-drug interaction. Our experimental results show that UniMAP significantly outperforms existing supervised and pre-training methods. We also conduct an ablation study on the effect of different losses, and observe an interesting complementary relation between global molecular-level pre-training loss and fine-grained local semantic representation loss. Furthermore, we conduct a visualization analysis on the output embeddings, and find that the clusters of both molecular and fragment representations demonstrate meaningful chemical properties. In addition, the learned attention weights of UniMAP show a clear fragment alignment pattern, validating the soundness of UniMAP.

Though this paper focuses on SMILES and graph data fusion, our proposed UniMAP is a modality-agnostic framework and therefore has the potential to be extended to a universal model, including various molecular data formats and downstream tasks, which could inspire diverse directions for future exploration.





\bibliographystyle{unsrt}
\bibliography{reference}

\appendix

\section{Open Access to Data and Code for Reproducibility} ~\label{opensource code}
The code is released publicly at \url{https://anonymous.4open.science/r/UniMAP-3CD7} for reproducibility.

\section{Experiments Details} \label{exp details}
The supplementary pre-training hyperparameters for UniMAP are outlined in Table~\ref{table:app setting pretrain}. In the context of fine-tuning, we furnish the specifics of the search space in Table~\ref{table:app setting molnet} for MoleculeNet grid search. The fine-tuning hyperparameters for tasks such as DDI prediction, and DTA prediction tasks (including DAVIS and KIBA) can be found in Table~\ref{table:app setting ddi dta}.

\begin{table}[h!]
    \caption{Hyperparameters for the UniMAP pre-training. }
    \label{table:app setting pretrain}
    \begin{center}
    \begin{footnotesize}
    \begin{tabular}{lc}
    \toprule
    Parameter & Value or description\\
     \midrule 
   Batch size &  256\\
Optimizer  & 	AdamW	\\
Adam betas & (0.9, 0.999)\\
Max Learning rate & 	5e-5	\\
Warm up steps & 	10000	\\
Learning rate decay policy	 & linear decay\\
 Training epochs & 40 \\
 \midrule 
 GNN encoder layers number & 3 \\
 GNN embedding size & 384 \\
 Transformer layers number & 12 \\
 Transformer embedding size & 768 \\
 \midrule 
token-level masking ratio & 0.2 \\
fragment-level masking ratio & 0.6 \\
temperature $\tau$ for contrastive loss & 0.05 \\
   \bottomrule
    \end{tabular}
    \end{footnotesize}
    \end{center}
    \vskip -0.1in
\end{table}

\begin{table}[h!]
\setlength{\tabcolsep}{4pt}
\caption{Search space for the tasks in MoleculeNet dataset}
\label{table:app setting molnet}
\begin{center}
\begin{small}
\begin{tabular}{ll}
\hline
Parameter&Search Space\\\hline
Learning rate&{[1e-6,1e-4]}\\
Batch size&\{16,32,64\}\\
Weight decay&[1e-7,1e-3]\\\hline
\end{tabular}
\end{small}
\end{center}
\vskip -0.1in
\end{table}

\begin{table}[h!]
    \caption{Hyperparameters for fine-tuning on DDI prediction task and DTA prediction tasks. }
    \label{table:app setting ddi dta}
    \begin{center}
    \begin{footnotesize}
    \begin{tabular}{lc|cc}
    \toprule
    Parameter & DDI & Davis & KIBA \\ 
     \midrule  
  Batch size &	32	& 64 & 64 \\
Learning rate	& 3e-5 & 2e-5 & 1e-4\\
Weight decay	& 1e-6	& 1e-6 & 1e-5	\\
   \bottomrule
    \end{tabular}
    \end{footnotesize}
    \end{center}
\end{table}

\section{Related Work}
Here we briefly introduce some typical molecular representation learning methods, including SMILES based methods, graph based methods, 3D based methods, and hybrid methods using multiple kinds of molecular input.

\subsection{SMILES based Methods}

Due to the sequential format of a SMILES string, NLP based pre-training methods~\cite{devlin-etal-2019-bert,danqi2019roberta,Zhu2020bert-nmt}, have been recently applied to SMILES and achieved competitive results. For example, SMILES-BERT~\cite{wang2019smilesbert} and ChemBERTa~\cite{chithrananda2020chemberta} use BERT and RoBERTa as the backbone to conduct MLM on SMILES, respectively. While SMILES Transformer~\cite{honda2019smilestrans} and X-MOL~\cite{xue2021xmol} leverage an encoder-decoder architecture to learn the representation by SMILES generation. Owing to the success of large-scale language models, these models have the ability to utilize large-scale molecular data for pre-training, e.g.~1.1 billion molecules in X-MOL. However, previous work~\cite{rong2020grover} has shown that these methods are limited in capturing molecular structure.

\subsection{Graph based Methods}
In general, graph based pre-training methods can be mainly classified into three categories: generative, predictive, and contrastive methods. The objective of generative methods is to reconstruct the original molecular graph or generate 1D sequence by 2D graph. For example, MGSSL~\cite{zhang2021mgssl} generates the motif tree of molecules using a pre-defined motif vocabulary, PanGu~\cite{lin2022pangu} takes 2D graph as input to generate 1D Self-Referencing Embedded Strings (SELFIES) ~\cite{krenn2020self} var a conditional variational autoencoder. Predictive methods construct context based on the graph structure and adopt context-aware masking for self-supervised learning. For example in ~\cite{hu2019strategies,rong2020grover}, they define neighbors in a graph as context and use the corresponding context on the molecular graph to predict the masked atoms/edges/motifs. While contrastive methods mainly apply contrastive learning on molecular graph. For example, MolCLR~\cite{wang2022molclr} and iMolCLR~\cite{wang2022imolclr} firstly transform graphs by atom/edge/sub-graph masking to construct positive and negative pairs, and utilize a constrastive loss. Besides, some other works use chemical knowledge for data augmentation in contrastive learning, e.g.~KCL~\cite{fang2022kcl} and MPG~\cite{li2021mpg}.

\subsection{3D based Methods}
Recently, several methods have been proposed to pre-train on 3D molecular conformations. For example, GraphMVP~\cite{liu2022graphmvp}, 3D-Infomax~\cite{stark20223d}, and GeomGCL~\cite{li2022geomgcl} adopt contrastive or generative learning framework between 2D graph and 3D input, to leverage 3D information to enhance the representations of 2D graph. GEM~\cite{fang2022geometry} develops a geometry learning task to predict the atom distances and angles from 3D conformation. Besides, some works such as Uni-Mol~\cite{zhou2023unimol}, Transformer-M~\cite{luo2022one} and 3D-EMGP~\cite{jiao2022energy}, propose to use 3D denoising methods to directly obtain the 3D molecular representation, and has achieved remarkable performances due to the proven equivalence between 3D denoising objective and learning a force field~\cite{zaidi2022pre}. To our understanding, representation using 3D data is definitely a promising direction. Still, 3D data are relatively not so large as compared with 1D or 2D data. Besides, what objective is the best to reflect the characteristics of 3D conformation is not clear yet~\cite{feng2024may, xia2022systematic}. Therefore, how to learn a reliable 3D conformation representation remains a challenging problem.


\subsection{Hybrid Methods}
Since a molecule can be represented as multiple forms, e.g.~SMILES, 2D graph, 3D conformations, IUPAC~\cite{panico1993iupac}, and cell-based microscopy images, some hybrid methods have been proposed to combine different forms to learn a unified representation. DMP~\cite{zhu2021dmp}, MM-Deacon~\cite{guo2022mmdeacon} and CLOOME~\cite{sanchez-fernandez2022cloome} combine 2D graph with SMILES, IUPAC with SMILES, and cell-based microscopy image with 2D graph, respectively. Moreover, MOCO~\cite{zhu2022improving} takes advantage of four molecular data forms containing SMILES, 2D graph, 3D conformations, and Morgan fingerprints for pre-training. However, these methods all use a two-stream or multi-stream model architecture, in which a separate encoder is first employed to obtain the representation of each data form, and then these representations are aligned with further loss, such as view consistency (in MOCO), multi-lingual contrastive alignment (in MM-Deacon), and InfoLOOB contrastive loss (in CLOOME). These two-stream models are good at learning the representation of each form, but miss the rich alignment information between different forms~\cite{huo2021wenlan}.

\section{Relations of UniMAP with Previous Work}
Our work presents the first single-stream hybrid model, to adapt to the corresponding relation between SMILES and graph. The basic idea of fine-grained alignment, though designed for SMILES and graph in our work, is also valid for combining other data forms, such as IUPAC and cell-based microscopy images. In the future, we will investigate how to design a general framework to integrate different data forms. 

Our work is also highly inspired by some recent vision-language multi-modality research, which can be categorized into two-stream and single-stream methods. According to the analysis in WenLan\cite{huo2021wenlan} and SemVLP\cite{li2021semvlp}, two-stream methods such as CLIP\cite{radford2021learning}, ALIGN\cite{jia2021scaling} and WenLan are good at capturing weak correlation between vision and language, while single-stream methods such as Oscar\cite{li2020oscar} and Uniter\cite{chen2020uniter} are more suitable to model strong correlation. From our study, SMILES and graph have a typical strong correlation, i.e.~SMILES provides detailed description of graph, just like the strong correlation between image and language in the image caption task. That is why we think a single-stream architecture is better for fusion between different data forms of a molecule.

\section{Drug-Drug Interaction Prediction} \label{ddi section}

\textbf{Data.} In addition to molecular property prediction and drug-target affinity prediction, we also conduct experiments on the drug-drug interaction (DDI) prediction task, which plays a crucial role in drug repositioning and virtual screening. Specifically, we experiment on the DrugBank Multi-Typed DDI dataset~\cite{ryu2018deepddi}, which contains 192,284 DDI pairs covering 86 interaction types. The objective is to predict the interaction type of each drug pair. The dataset is split to train, validation, and test set with the ratio 3:1:1. We randomly split the data 3 times with different seeds and report the mean value of evaluation metrics, including accuracy, precision, recall, and F1-score as in~\cite{chen2021muffin}. 

\textbf{Baseline Models.} We compare UniMAP with different kinds of baselines, including deep learning method DeepDDI~\cite{ryu2018deepddi}, graph embedding methods DeepWalk~\cite{perozzi2014deepwalk} and LINE~\cite{tang2015line}, and some recent pre-training methods such as MUFFIN~\cite{chen2021muffin}, MUFFIN\_KG~\cite{chen2021muffin}, X-MOL~\cite{xue2021xmol}, PanGu~\cite{lin2022pangu} 
and REMO~\cite{tang2024contextual}. Please note that we use MUFFIN\_KG to refer to the MUFFIN variant with an extra knowledge graph branch.


\begin{table}[h!]
\caption{Results for DDI task. We report the accuracy, precision, recall and F1-score, the best and second best results are marked bold and \underline{underlined}, respectively.}
\centering
\begin{tabular}{lllll}
\hline
\textbf{Method}        & \multicolumn{1}{c}{\textbf{Accuracy}}       & \textbf{Precision}      & \textbf{Recall}         & \textbf{F1-score}             \\ \hline
DeepDDI   & 0.877                     & 0.799 & 0.759 & 0.766 \\
DeepWalk   & 0.800                     & 0.822 & 0.710 & 0.747 \\
LINE      & 0.751                     & 0.687 & 0.545 & 0.580 \\
MUFFIN & 0.939                     & 0.926 & 0.908 & 0.911 \\
MUFFIN\_KG  & \underline {0.965}                              & \underline{0.957} & \textbf{0.948} & \underline{0.950} \\
X-MOL      & 0.952 & -     & -     & -     \\
PanGu      & 0.957 & -     & -     & -     \\ 
REMO & 0.953 & 0.932 &0.932 &0.928 \\
\hline
\textbf{UniMAP} & \textbf{0.972} & \textbf{0.958}    & \underline {0.943}    & \textbf {0.952}    \\
\hline
\end{tabular}%
\label{ddi}
\end{table}

\textbf{Fine-Tuning UniMAP.}
For the downstream fine-tuning of UniMAP, we first concatenate the representation $x_{cls}$ of the two paired drugs as the input, and then feed it to a two-layer MLP for multi-class prediction.

\textbf{Results.} The experimental results are shown in Table \ref{ddi}. Here we report four metrics including accuracy, precision, recall, and F1-score, where X-MOL and PanGu only show the accuracy result in their original paper. We can see that all pre-training methods (MUFFIN, MUFFIN\_KG, X-MOL, Pangu, and UniMAP) outperform deep learning method (DeepDDI) and graph embedding methods (DeepWalk and LINE), demonstrating the benefit of pre-training. More importantly, UniMAP consistently outperforms the pre-training methods in terms of all four metrics, except for the recall compared with MUFFIN\_KG. This may be because MUFFIN\_KG utilizes extra knowledge graph information, which is not considered in both UniMAP and other pre-training baselines. Therefore, UniMAP shows better DDI prediction results by leveraging multi-modality pre-training, compared with single-modality pre-training methods, such as MUFFIN, X-MOL, and PanGu.



\section{Regression Tasks in MoleculeNet}

\begin{table}[h!]
    \centering
    \caption{Results in terms of mean and std for the 4 regression tasks. We report RMSE scores, the best and second best results are marked as \textbf{bold} and \underline{underlined}, respectively.}
    \scalebox{1.0}{
    \begin{tabular}{lllll}
    \hline
        \textbf{Method} & \textbf{ESOL} & \textbf{Lipo} & \textbf{Malaria} & \textbf{CEP} \\ \hline
        GROVER(base) & \underline{0.904(0.061)} & 0.815(0.012) & 1.092(0.007) & 1.339(0.005) \\ 
        GROVER(large) & 0.951(0.027) & 0.803(0.038) & 1.082(0.009) & 1.353(0.004) \\ 
        GraphMVP & 1.064(0.045) & 0.691(0.013) & 1.106(0.013) & 1.228(0.001) \\ 
        Mole-BERT & 1.015(0.030) & \underline{0.676(0.017)} & \underline{1.074(0.009)} & 1.232(0.009) \\ 
        MOCO &0.984(0.034) & 0.707(0.001) &1.093(0.009) & \textbf{1.101(0.007)} \\ \hline
        UniMAP & \textbf{0.861(0.010)} & \textbf{0.664(0.023)} & \textbf{1.043(0.007)} & \underline{1.128(0.007)} \\ \hline
    \end{tabular}
    }
    \label{molnetreg}
\end{table}
Beyond the classification tasks, we also verify our method's effectiveness on 4 regression tasks: ESOL~\cite{delaney2004esol} contains regression labels measuring the molecular water solubility. Lipophilicity(Lipo), which is a subset of ChEMBL~\cite{gaulton2012chembl}, consists of experimental results of octanol/water distribution coefficient. CEP records the organic photovoltaic efficiency of molecules selected from the Havard Clean Energy Project(CEP)~\cite{hachmann2011harvard}. Malaria~\cite{gamo2010thousands} takes drug efficacy against malaria, which is a type of disease caused by parasites, as regression target. We choose the top 4 methods based on the averaged classification results in Table~\ref{molnetcls} for comparison. That is to say, MOCO, Mole-BERT, GROVER, and GraphMVP are treated as our baseline methods. Similarly, a two-layer MLP is adopted to obtain the final regression value of UniMAP, supervised by a corresponding regression loss, i.e.~MAE. RMSE is utilized as the evaluation metric. From the results in Table~\ref{molnetreg}, we can see that UniMAP performs the best on three tasks, and second best on the other one.

\section{Fragment Alignment Pattern}

\begin{figure*}[htb]
\centering
\subfigure[]{
\label{attention.sub.a}
\includegraphics[scale=0.3]{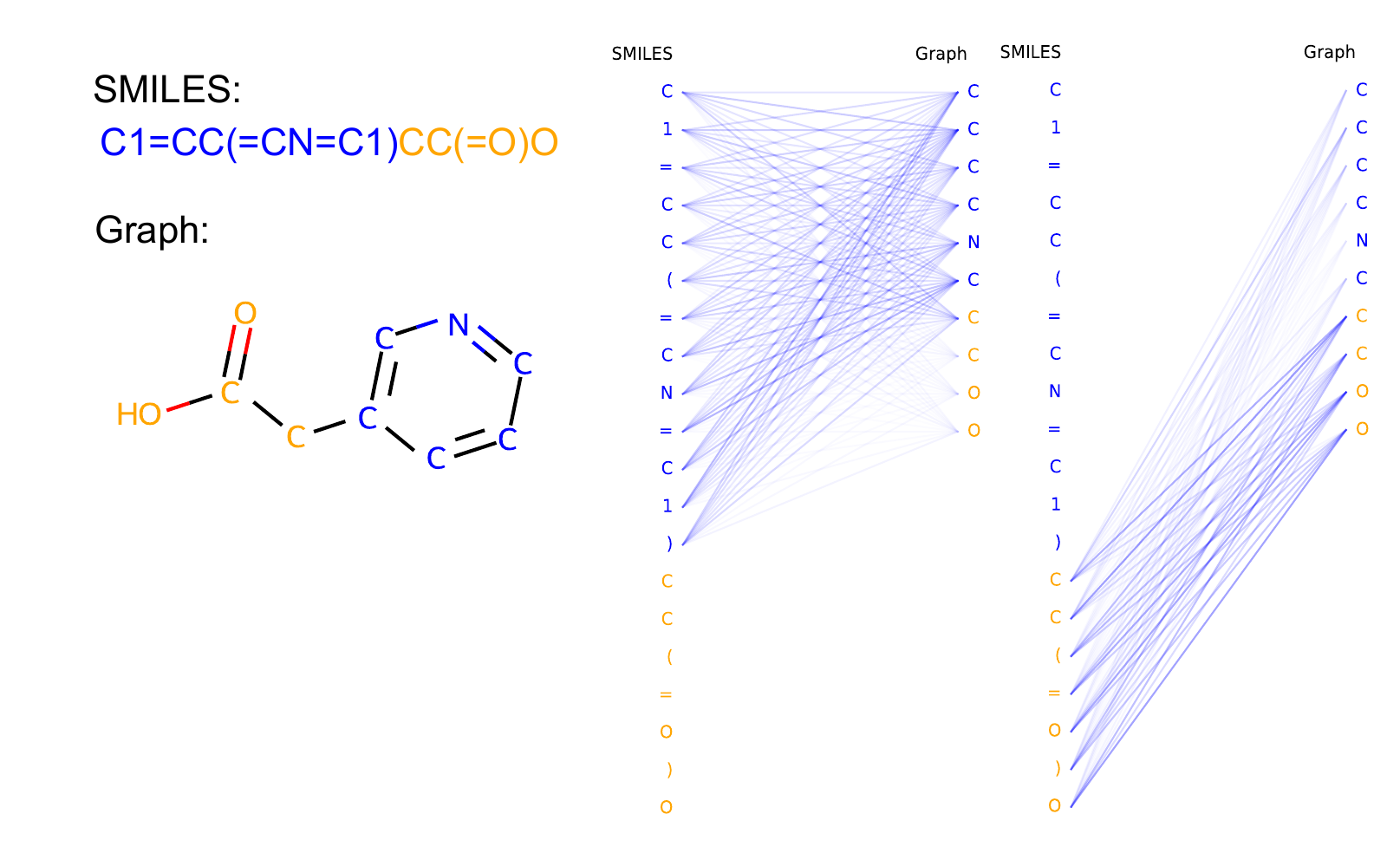}}
\subfigure[]{
\label{attention.sub.b}
\includegraphics[scale=0.52]{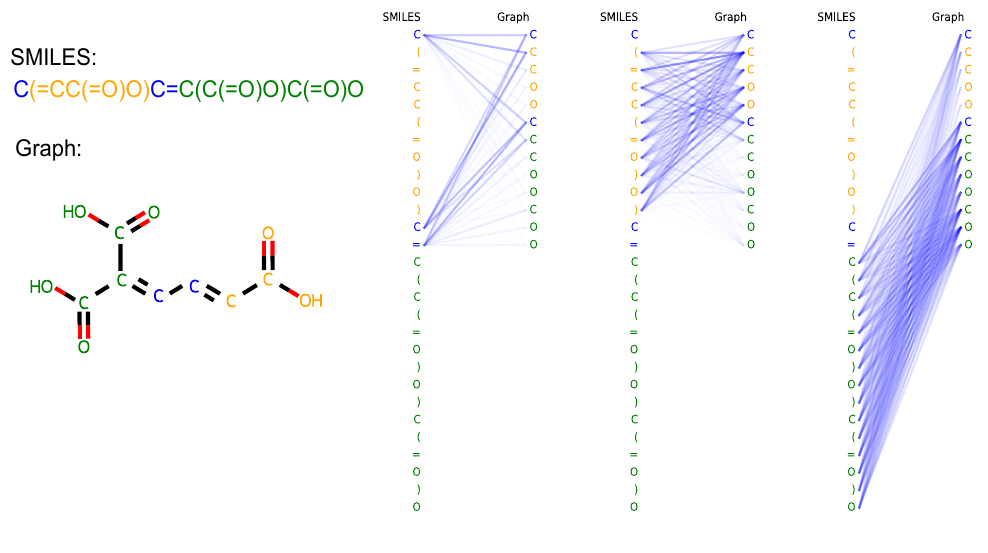}}
\subfigure[]{
\label{attention.sub.c}
\includegraphics[scale=0.6]{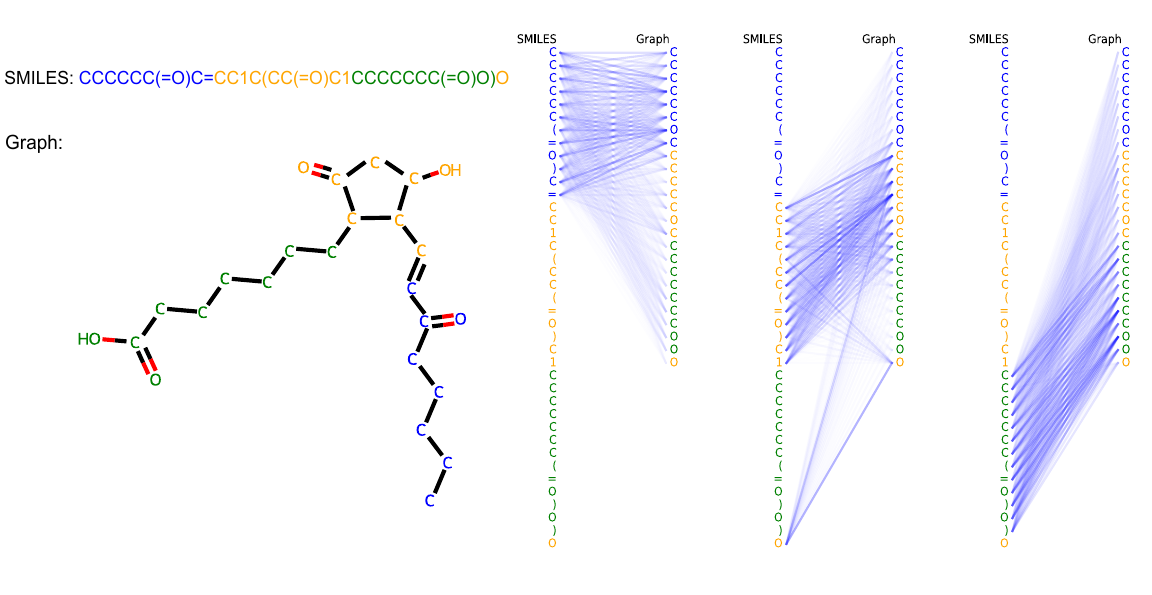}}
\caption{Visualization of attentions from SMILES to graph.}
\label{attention}
\end{figure*}

Inspired by the analysis in NLP~\cite{clark2019does} that higher layer attention weights usually demonstrate some learned patterns, we visualize the learned attention weights at the 8th layer. Interestingly, we find that the learned model demonstrates a clear fragment alignment pattern. We give three example molecules as case study, as shown in Figure \ref{attention}. Given the SMILES and graph representation of a molecule, multiple fragments are obtained, and labeled with different colors, and the attention weights between different SMILES tokens and graph atoms are presented to lines with different color depths, where deeper color means higher attention weight. Clearly, the larger attention weights are obtained between SMILES tokens and graph atoms within the same corresponding fragment, validating the soundness of UniMAP.

The alignment of fragments not only facilitates the seamless fusion of SMILES and graphs in a finer-grained manner but also enables the model to highlight the differences among fragments, thereby enhancing the comprehension of molecular properties. Although SMILES better captures global contextual information~\cite{zhu2021dmp}, it lacks the ability to model meaningful substructures like fragments. For instance, in Figure 1, the adjacency of the yellow fragment in the SMILES is interrupted by another green-colored fragment. Through cross modality fragment alignment,  Through cross-modality fragment alignment, the model successfully captures such fine-grained information via learned attention weights.
Alignment of fragments helps the model better comprehend the molecule by decomposition into fragments, When examining and designing molecules, trained chemists intuitively decompose them into fragments, based on several key factors such as synthetic availability, chemical properties, and biological functionalities~\cite{erlanson2004fragment}. Therefore, fragment decomposition is essential for molecular representation learning, as it captures local information such as functional groups and simplifies molecule complexity for machine learning models.

In particular, regarding the (a) sub-figure in Figure~\ref{attention}, UniMAP outlines the yellow fragment known as carboxymethyl and the blue fragment known as pyridyl. Carboxymethyl is a well-known acidic fragment. Its presence often contributes to a molecule's acidic character. Conversely, pyridyl which can readily accept protons is a recognized basic fragment. Additionally, this distinction in fragments also translates to different binding mechanisms. The carboxymethyl group, with its negative charge at physiological pH, can form strong ionic bonds with positively charged residues in the binding site, and it can act as both a hydrogen bond donor (due to the hydroxyl group) and acceptor (due to the carbonyl oxygen), further stabilizing the complex through forming potential hydrogen bonds. In contrast, the pyridyl can provide $\pi$-stacking interactions due to its aromaticity.

Finally, we re-run the code of MOCO on the three molecules in Figure~\ref{attention}. As the MOCO aligns the global representation of each modality by contrastive learning, we cannot provide the visualization of attentions. Instead, we calculate the cosine similarity between the corresponding fragments in SMILES and graph, the fragment embeddings are average pooling from the corresponding tokens/atoms embeddings from 1D/2D network. The results are presented in Table~\ref{tab:moco_fragsim}.

\begin{table}[h]
\centering
\begin{tabular}{cccc}
\hline
Molecule & Fragment1 & Fragment2 & Fragment3 \\ \hline
 a & 0.0737 & 0.0482 & - \\
 b & -0.0340 & 0.0258 & -0.0177 \\
 c & 0.0254 & 0.0227 & -0.0203 \\ \hline
\end{tabular}
\caption{The cosine similarity of corresponding fragments between graph and SMILES on the three molecules depicted in Figure~\ref{attention}}
\label{tab:moco_fragsim}
\end{table}

As evident from Table~\ref{tab:moco_fragsim}, the low similarity scores for fragments indicate it's difficult for single-stream methods like MOCO to capture fragment-level semantic information due to the lack of fine-grained interaction and fragment-level alignment supervision.


\section{Feature Visualization on ClinTox Task}

\begin{figure*}[t]
\centering
\includegraphics[scale=0.22]{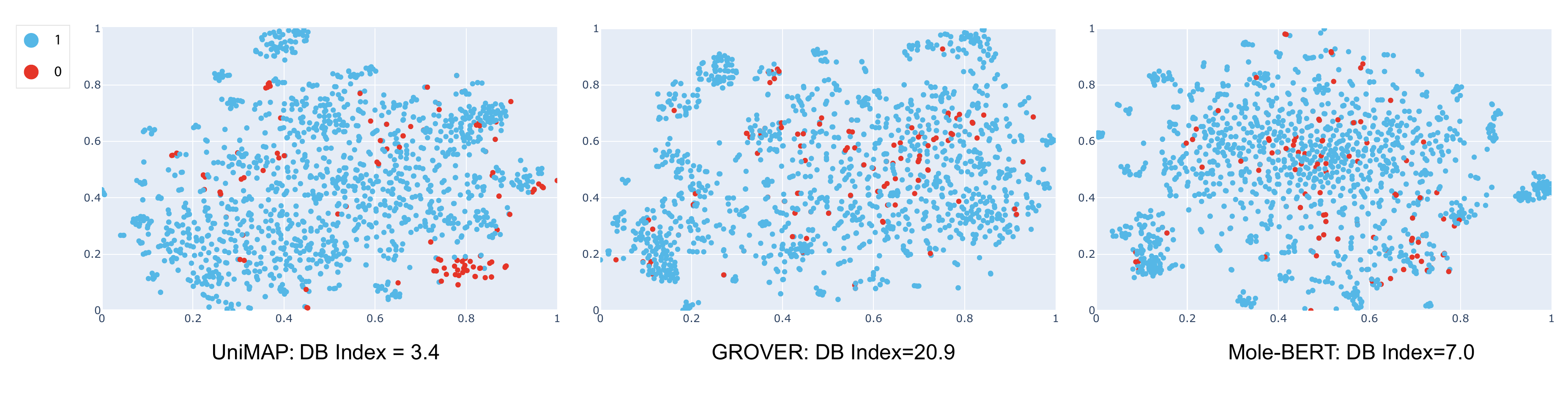}
\vspace{-0.2cm}\caption{The unfine-tuned feature visualization on unbalanced ClinTox dataset, where distinct colors indicate the binary label of each sample. UniMAP exhibits the best clustering results, as evidenced by the smallest DB Index. This aligns with its superior performance after fine-tuning present in Table~\ref{molnetcls}.}
\label{clintox_cluster}
\end{figure*}

From the results presented in Table~\ref{molnetcls}, it is evident that UniMAP significantly outperforms the second-best baseline (GROVER) by a substantial margin (98.3 vs. 81.7) on the ClinTox task. To delve into the reasons behind UniMAP's superiority, we extracted the unfine-tuned embeddings of three pretraining methods—UniMAP, Grover, and Mole-BERT. Subsequently, we employed T-SNE to visualize the clustering results, directly illustrating the effectiveness of the pretraining process. The clustering results are illustrated in Figure~\ref{clintox_cluster}, where distinct colors represent different binary labels, namely 0 or 1. Additionally, to provide a quantitative evaluation of the clustering performance, we introduce the Davies-Bouldin Index (DB Index)~\cite{davies1979cluster}. A lower DB Index value signifies superior clustering performance.
Upon examining Figure~\ref{clintox_cluster}, it becomes evident that the labels are highly unbalanced in the ClinTox dataset, as evidenced by the scarcity of red points. This imbalance poses a challenge for clustering. While the red points appear scattered in the case of GROVER and Mole-BERT, UniMAP effectively aggregates the limited red points into a meaningful cluster positioned in the bottom right of the figure. This capability is advantageous for the subsequent fine-tuning process, elucidating the superior performance of UniMAP.

\section{The Influence of Different Pooling Methods for Obtaining Fragment Embeddings}
We employ different pooling techniques for obtaining fragment-level representations of different modalities: multi-head attention with average pooling for SMILES, and average pooling for graphs. Our motivation here lies in the desire to employ various pooling methods tailored to different modality preferences. We supplement additional ablation experiments to verify the effect of different pooling methods. In particular, we add a setting that uses average pooling for both SMILES and graph and show the performance on MoleculeNet in Table~\ref{tab:pooling_abi}, the new setting is denoted as `Both Avg'. All settings only use 1M pre-training data to save time and computational cost.

\begin{table*}[t]
\centering
\scalebox{0.8}{
\begin{tabular}{lcccccccc}
\hline
Method & BBBP & Sider & ClinTox & Bace & Tox21 & Toxcast & MUV & HIV \\
\hline
UniMAP & \textbf{76.1(0.4)} & \textbf{65.8(0.1)} & 
\textbf{98.6(0.2)} & 80.9(0.3) & \textbf{76.3(0.6)} & 
\textbf{60.1(0.4)} & \textbf{72.6(0.3)} & \textbf{77.1(0.4)} \\
Both Avg & 73.7(1.8) & 63.5(1.3) & 97.2(1.2) & 
\textbf{82.9(0.7)} & 75.3(1.1) & 59.0(0.7) & 71.9(0.2) & 76.6(0.1) \\
\hline
\end{tabular}
}
\caption{Performance of different pooling methods for obtaining fragment embeddings on MoleculeNet.}
\label{tab:pooling_abi}
\end{table*}

As we can see from Table~\ref{tab:pooling_abi}, the performance of setting `Both Avg' overall is inferior to the baseline setting. Since the fragment pooling operation only affects fragment-level alignment, we examine the loss curves and observe that the FLA (Fragment-Level Alignment) loss does not converge to the level of the default setting, we conjecture this discrepancy may contribute to the degradation of performance.

\section{More Discussions about the Complementary between SMILES and Graph}

To better comprehend the complementary between SMILES and Graph in the single-stream approach.  We make modifications to our designation to pre-train separate encoders which take a single modality as input with UniMAP strategies. In particular, we replace the shared Transformer with separate encoders: one for encoding SMILES and another for encoding the graph. Additionally, we introduce a 2-layer multi-head attention module for later fusion of the embeddings from these separate encoders. The losses of UniMAP remained unchanged throughout this adaptation. Each setting is pre-trained on 1M molecules to save time. Then, we evaluate the performance of two single-modal models, utilizing either SMILES or graph as input, on the MoleculeNet dataset. The results are presented in Table~\ref{tab:complementary}

\begin{table*}[t]
\centering
\caption{Comparison of the performance between a single-stream model and separate single-modality models on MoleculeNet.}
\label{tab:complementary}
\scalebox{0.8}{
\begin{tabular}{lcccccccc}
\hline
Method      & BBBP           & Sider          & ClionTox       & Bace           & Tox21          & Toxcast        & MUV            & HIV            \\ \hline
UniMAP      & \textbf{76.1(0.4)}      & \textbf{65.8(0.1)}      & \textbf{98.6(0.2)}      & 80.9(0.3)      & \textbf{76.3(0.6)}      & 60.1(0.4)      & \textbf{72.6(0.3)}      & \textbf{77.1(0.4)}      \\
SMILES Only & 74.1(0.8)      & 64.2(0.6)      & 96.1(0.2)      & \textbf{81.2(0.2)}      & 73.6(1.3)      & 59.0(2.3)      & 71.4(1.2)      & 76.4(0.5)      \\
Graph Only  & 71.3(1.4)      & 62.9(0.7)      & 85.6(0.2)      & 72.1(1.2)      & 73.9(0.3)      & \textbf{60.9(0.7)}      & 69.8(0.4)      & 76.8(0.3)      \\ \hline
\end{tabular}}
\end{table*}

Table~\ref{tab:complementary} shows that the original setting which uses the single-stream model achieves better performance above the settings of separated encoders, indicating the single-stream model better captures the complementary of both input modalities.

\begin{figure*}[t]
\centering
\includegraphics[scale=0.22]{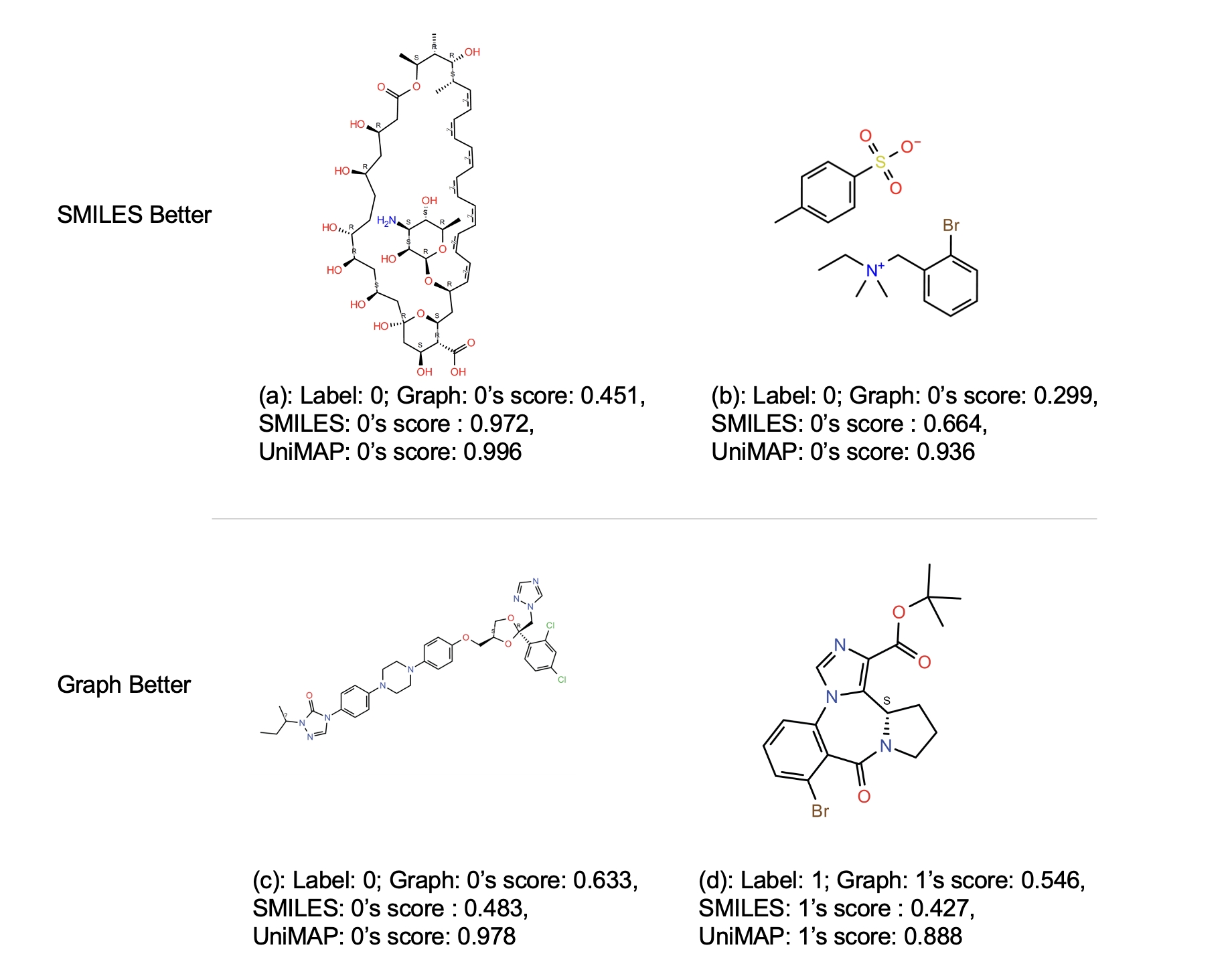}
\vspace{-0.2cm}\caption{Four typical molecules from BBBP task to illustrate the complementary between SMILES and graph. The first row including (a) and (b) demonstrates cases where the SMILES encoder outperforms the graph encoder, while in the second row including (b) and (c), the opposite is observed. The UniMAP which takes both as input to a single-stream model achieves accurate predictions for all four molecules. }
\label{bbbp_cases}
\end{figure*}

To better elucidate our conjecture, we delve into the BBBP task, wherein the single-stream model exhibits a significant performance boost. Leveraging inductive bias, we analyze typical samples along with their corresponding predictions generated by different models. Specifically, we select four molecules and visually depict the labels and predictions of each model in Figure~\ref{bbbp_cases}

In the first two cases (a) and (b), SMILES outperforms Graph. The first molecule features a large ring and multiple chiral centers. SMILES, being a sequential representation, excels in modeling long-distance interactions~\cite{zhu2021dmp} and exhibits greater sensitivity in chirality predictions due to the explicit representation of chirality using the '@' character~\cite{zhu2022improving}. Consequently, it predicts such molecules more accurately. The second molecule (b) is an ionic compound lacking covalent connections between its ionic moieties. In 2D graphs, different ionic moieties are often depicted as independent graphs, each illustrating the connectivity of atoms within a single ionic moiety. However, SMILES connects the SMILES representations of different ionic moieties using the symbol ".", creating a contiguous sequence that models fine-grained interactions, helping the model comprehend the overall contextual information of such molecules.

In the last two cases (c) and (d), Graph outperforms SMILES. The first molecule contains many rings, which increase rigidity and make it hard to pass through the blood-brain barrier. The second molecule contains a benzodiazepine core fragment, a meaningful substructure recognized for its anti-anxiety and anti-seizure properties, indicating it can permeability through the blood-brain barrier~\cite{haefely1992partial}. 2D Graphs outperform in capturing locally significant structures and topological information of molecules, including ring count and modeling substructures. Conversely, in SMILES, such fragments are often non-adjacent and interpreted by other fragments (e.g., the blue fragment in Figure 5b and the yellow fragment in Figure 5c), posing challenges in outlining local topological structures.

Overall, UniMAP consistently outperforms single-modality based models in all four molecules, indicating the effectiveness of our single-stream approach. Properties like BBBP are influenced by various factors such as chirality, rigidity, and specific fragments. Simultaneously inputting SMILES and 2D Graph into a single-stream model enhances the depiction of these aspects, thereby improving overall performance.

\section{Comparison between Single-stream Methods and Two-stream Methods in Molecular Representation Learning Domain}

Similar to the vision-language domain, single-stream approaches such as Transformer-M, MoleBLEND, and BindNet~\cite{feng2023protein} excel in capturing fine-grained correlations and achieving higher performance. However, they require pairwise data and involve additional computational costs due to cross-modality fine-grained interaction.

On the other hand, two-stream methods such as GraphMVP, 3D-Infomax, and DrugCLIP~\cite{gao2024drugclip} are designed to model weaker correlations, offering flexibility in network design. These methods are particularly suitable for retrieval scenarios, as feature calculation can be conducted offline. For instance, the two-stream method DrugCLIP significantly accelerates the speed of virtual screening.

Transformer-M and MoleBLEND, as other fine-grained pre-training methods, attempt to fuse 3D distance information and 2D topological paths into the attention bias of the transformer. We provide a comparison with MoleBLEND in Table~\ref{tab:compare_molblend} on MoleculeNet. Our superior performance aligns with our inductive bias that molecular fragments are more relevant to biological properties and thus more valuable to be considered in fine-grained approaches.

\begin{table*}[t]
\centering
\scalebox{0.8}{
\begin{tabular}{lccccccccc}
\hline
Method & BBBP & Tox21 & MUV & BACE & ToxCast & SIDER & ClinTox & HIV & Avg.\\
\hline
MoleBLEND & 73.0(0.8) & \textbf{77.8(0.8)} &  77.2(2.3)  & \textbf{83.7(1.4)} & 66.1(0.0) & 64.9(0.3) & 87.6(0.7) &79.0(0.8)& 76.2 \\
UniMAP & \textbf{75.6(1.8)} & 77.2(0.9) & \textbf{ 78.9(1.4)} & 83.6(1.0) & \textbf{67.0(0.4)} & \textbf{66.6(0.8)} & \textbf{98.3(1.7)} & \textbf{80.1(0.7)} & \textbf{78.4} \\
\hline
\end{tabular}
}
\caption{Comparision of the performance between UniMAP and MoleBLEND on MoleculeNet.}
\label{tab:compare_molblend}
\end{table*}

\section{Limitation} \label{limitation}
UniMAP focuses on the fusion of 1D SMILES and 2D molecular graphs, without involving 3D modality. The learning of 3D representations needs input data with 3D conformations. However, many downstream datasets, particularly those we focus on at UniMAP, such as MoleculeNet, Davis, and KIBA,  don't originally provide the 3D structures. Consequently, training a 3D-based encoder needs additional computation costs to generate the 3D conformations and might restrict the scope of applicability of methods. However, leveraging 3D conformation offers additional structural information compared to 2D or 1D representations, the motivation that captures fine-grained semantics between different modalities is compatible with 3D,  thus,  the integration of 3D data into UniMAP is worth exploring for future work.



\newpage
\section*{NeurIPS Paper Checklist}

\begin{enumerate}

\item {\bf Claims}
    \item[] Question: Do the main claims made in the abstract and introduction accurately reflect the paper's contributions and scope?
    \item[] Answer: \answerYes{} 
    \item[] Justification: 
    The main claims made in the abstract and introduction accurately reflect the paper's contributions and scope.
    \item[] Guidelines:
    \begin{itemize}
        \item The answer NA means that the abstract and introduction do not include the claims made in the paper.
        \item The abstract and/or introduction should clearly state the claims made, including the contributions made in the paper and important assumptions and limitations. A No or NA answer to this question will not be perceived well by the reviewers. 
        \item The claims made should match theoretical and experimental results, and reflect how much the results can be expected to generalize to other settings. 
        \item It is fine to include aspirational goals as motivation as long as it is clear that these goals are not attained by the paper. 
    \end{itemize}

\item {\bf Limitations}
    \item[] Question: Does the paper discuss the limitations of the work performed by the authors?
    \item[] Answer: \answerYes{} 
    \item[] Justification: 
    We discuss the limitation of UniMAP at appendix~\ref{limitation}
    \item[] Guidelines:
    \begin{itemize}
        \item The answer NA means that the paper has no limitation while the answer No means that the paper has limitations, but those are not discussed in the paper. 
        \item The authors are encouraged to create a separate "Limitations" section in their paper.
        \item The paper should point out any strong assumptions and how robust the results are to violations of these assumptions (e.g., independence assumptions, noiseless settings, model well-specification, asymptotic approximations only holding locally). The authors should reflect on how these assumptions might be violated in practice and what the implications would be.
        \item The authors should reflect on the scope of the claims made, e.g., if the approach was only tested on a few datasets or with a few runs. In general, empirical results often depend on implicit assumptions, which should be articulated.
        \item The authors should reflect on the factors that influence the performance of the approach. For example, a facial recognition algorithm may perform poorly when image resolution is low or images are taken in low lighting. Or a speech-to-text system might not be used reliably to provide closed captions for online lectures because it fails to handle technical jargon.
        \item The authors should discuss the computational efficiency of the proposed algorithms and how they scale with dataset size.
        \item If applicable, the authors should discuss possible limitations of their approach to address problems of privacy and fairness.
        \item While the authors might fear that complete honesty about limitations might be used by reviewers as grounds for rejection, a worse outcome might be that reviewers discover limitations that aren't acknowledged in the paper. The authors should use their best judgment and recognize that individual actions in favor of transparency play an important role in developing norms that preserve the integrity of the community. Reviewers will be specifically instructed to not penalize honesty concerning limitations.
    \end{itemize}

\item {\bf Theory Assumptions and Proofs}
    \item[] Question: For each theoretical result, does the paper provide the full set of assumptions and a complete (and correct) proof?
    \item[] Answer: \answerNA{} 
    \item[] Justification: Our paper does not include theoretical results.
    \item[] Guidelines:
    \begin{itemize}
        \item The answer NA means that the paper does not include theoretical results. 
        \item All the theorems, formulas, and proofs in the paper should be numbered and cross-referenced.
        \item All assumptions should be clearly stated or referenced in the statement of any theorems.
        \item The proofs can either appear in the main paper or the supplemental material, but if they appear in the supplemental material, the authors are encouraged to provide a short proof sketch to provide intuition. 
        \item Inversely, any informal proof provided in the core of the paper should be complemented by formal proofs provided in appendix or supplemental material.
        \item Theorems and Lemmas that the proof relies upon should be properly referenced. 
    \end{itemize}

    \item {\bf Experimental Result Reproducibility}
    \item[] Question: Does the paper fully disclose all the information needed to reproduce the main experimental results of the paper to the extent that it affects the main claims and/or conclusions of the paper (regardless of whether the code and data are provided or not)?
    \item[] Answer: \answerYes{} 
    \item[] Justification: 
    We have provided detailed descriptions of the experiments along with open-source code to reproduce our results.
    \item[] Guidelines:
    \begin{itemize}
        \item The answer NA means that the paper does not include experiments.
        \item If the paper includes experiments, a No answer to this question will not be perceived well by the reviewers: Making the paper reproducible is important, regardless of whether the code and data are provided or not.
        \item If the contribution is a dataset and/or model, the authors should describe the steps taken to make their results reproducible or verifiable. 
        \item Depending on the contribution, reproducibility can be accomplished in various ways. For example, if the contribution is a novel architecture, describing the architecture fully might suffice, or if the contribution is a specific model and empirical evaluation, it may be necessary to either make it possible for others to replicate the model with the same dataset, or provide access to the model. In general. releasing code and data is often one good way to accomplish this, but reproducibility can also be provided via detailed instructions for how to replicate the results, access to a hosted model (e.g., in the case of a large language model), releasing of a model checkpoint, or other means that are appropriate to the research performed.
        \item While NeurIPS does not require releasing code, the conference does require all submissions to provide some reasonable avenue for reproducibility, which may depend on the nature of the contribution. For example
        \begin{enumerate}
            \item If the contribution is primarily a new algorithm, the paper should make it clear how to reproduce that algorithm.
            \item If the contribution is primarily a new model architecture, the paper should describe the architecture clearly and fully.
            \item If the contribution is a new model (e.g., a large language model), then there should either be a way to access this model for reproducing the results or a way to reproduce the model (e.g., with an open-source dataset or instructions for how to construct the dataset).
            \item We recognize that reproducibility may be tricky in some cases, in which case authors are welcome to describe the particular way they provide for reproducibility. In the case of closed-source models, it may be that access to the model is limited in some way (e.g., to registered users), but it should be possible for other researchers to have some path to reproducing or verifying the results.
        \end{enumerate}
    \end{itemize}

\item {\bf Open access to data and code}
    \item[] Question: Does the paper provide open access to the data and code, with sufficient instructions to faithfully reproduce the main experimental results, as described in supplemental material?
    \item[] Answer: \answerYes{} 
    \item[] Justification: We have provided the open-source code and data in appendix~\ref{opensource code}
    \item[] Guidelines:
    \begin{itemize}
        \item The answer NA means that paper does not include experiments requiring code.
        \item Please see the NeurIPS code and data submission guidelines (\url{https://nips.cc/public/guides/CodeSubmissionPolicy}) for more details.
        \item While we encourage the release of code and data, we understand that this might not be possible, so “No” is an acceptable answer. Papers cannot be rejected simply for not including code, unless this is central to the contribution (e.g., for a new open-source benchmark).
        \item The instructions should contain the exact command and environment needed to run to reproduce the results. See the NeurIPS code and data submission guidelines (\url{https://nips.cc/public/guides/CodeSubmissionPolicy}) for more details.
        \item The authors should provide instructions on data access and preparation, including how to access the raw data, preprocessed data, intermediate data, and generated data, etc.
        \item The authors should provide scripts to reproduce all experimental results for the new proposed method and baselines. If only a subset of experiments are reproducible, they should state which ones are omitted from the script and why.
        \item At submission time, to preserve anonymity, the authors should release anonymized versions (if applicable).
        \item Providing as much information as possible in supplemental material (appended to the paper) is recommended, but including URLs to data and code is permitted.
    \end{itemize}

\item {\bf Experimental Setting/Details}
    \item[] Question: Does the paper specify all the training and test details (e.g., data splits, hyperparameters, how they were chosen, type of optimizer, etc.) necessary to understand the results?
    \item[] Answer: \answerYes{} 
    \item[] Justification: We have specified the training and test details in section~\ref{experiments} and appendix~\ref{exp details}.
    \item[] Guidelines:
    \begin{itemize}
        \item The answer NA means that the paper does not include experiments.
        \item The experimental setting should be presented in the core of the paper to a level of detail that is necessary to appreciate the results and make sense of them.
        \item The full details can be provided either with the code, in appendix, or as supplemental material.
    \end{itemize}

\item {\bf Experiment Statistical Significance}
    \item[] Question: Does the paper report error bars suitably and correctly defined or other appropriate information about the statistical significance of the experiments?
    \item[] Answer: \answerYes{} 
    \item[] Justification: We have reported the mean and standard deviation values for three different random seeds in the MoleculeNet experiments.
    \item[] Guidelines:
    \begin{itemize}
        \item The answer NA means that the paper does not include experiments.
        \item The authors should answer "Yes" if the results are accompanied by error bars, confidence intervals, or statistical significance tests, at least for the experiments that support the main claims of the paper.
        \item The factors of variability that the error bars are capturing should be clearly stated (for example, train/test split, initialization, random drawing of some parameter, or overall run with given experimental conditions).
        \item The method for calculating the error bars should be explained (closed form formula, call to a library function, bootstrap, etc.)
        \item The assumptions made should be given (e.g., Normally distributed errors).
        \item It should be clear whether the error bar is the standard deviation or the standard error of the mean.
        \item It is OK to report 1-sigma error bars, but one should state it. The authors should preferably report a 2-sigma error bar than state that they have a 96\% CI, if the hypothesis of Normality of errors is not verified.
        \item For asymmetric distributions, the authors should be careful not to show in tables or figures symmetric error bars that would yield results that are out of range (e.g. negative error rates).
        \item If error bars are reported in tables or plots, The authors should explain in the text how they were calculated and reference the corresponding figures or tables in the text.
    \end{itemize}

\item {\bf Experiments Compute Resources}
    \item[] Question: For each experiment, does the paper provide sufficient information on the computer resources (type of compute workers, memory, time of execution) needed to reproduce the experiments?
    \item[] Answer: \answerYes{} 
    \item[] Justification: The details about the computer resource are provided in section~\ref{experiments}.
    \item[] Guidelines:
    \begin{itemize}
        \item The answer NA means that the paper does not include experiments.
        \item The paper should indicate the type of compute workers CPU or GPU, internal cluster, or cloud provider, including relevant memory and storage.
        \item The paper should provide the amount of compute required for each of the individual experimental runs as well as estimate the total compute. 
        \item The paper should disclose whether the full research project required more compute than the experiments reported in the paper (e.g., preliminary or failed experiments that didn't make it into the paper). 
    \end{itemize}
    
\item {\bf Code Of Ethics}
    \item[] Question: Does the research conducted in the paper conform, in every respect, with the NeurIPS Code of Ethics \url{https://neurips.cc/public/EthicsGuidelines}?
    \item[] Answer: \answerYes{} 
    \item[] Justification: The research conducted in the paper conforms with the NeurIPS Code of Ethics.
    \item[] Guidelines:
    \begin{itemize}
        \item The answer NA means that the authors have not reviewed the NeurIPS Code of Ethics.
        \item If the authors answer No, they should explain the special circumstances that require a deviation from the Code of Ethics.
        \item The authors should make sure to preserve anonymity (e.g., if there is a special consideration due to laws or regulations in their jurisdiction).
    \end{itemize}

\item {\bf Broader Impacts}
    \item[] Question: Does the paper discuss both potential positive societal impacts and negative societal impacts of the work performed?
    \item[] Answer: \answerNA{} 
    \item[] Justification: There is no societal impact of the work performed. 
    \item[] Guidelines:
    \begin{itemize}
        \item The answer NA means that there is no societal impact of the work performed.
        \item If the authors answer NA or No, they should explain why their work has no societal impact or why the paper does not address societal impact.
        \item Examples of negative societal impacts include potential malicious or unintended uses (e.g., disinformation, generating fake profiles, surveillance), fairness considerations (e.g., deployment of technologies that could make decisions that unfairly impact specific groups), privacy considerations, and security considerations.
        \item The conference expects that many papers will be foundational research and not tied to particular applications, let alone deployments. However, if there is a direct path to any negative applications, the authors should point it out. For example, it is legitimate to point out that an improvement in the quality of generative models could be used to generate deepfakes for disinformation. On the other hand, it is not needed to point out that a generic algorithm for optimizing neural networks could enable people to train models that generate Deepfakes faster.
        \item The authors should consider possible harms that could arise when the technology is being used as intended and functioning correctly, harms that could arise when the technology is being used as intended but gives incorrect results, and harms following from (intentional or unintentional) misuse of the technology.
        \item If there are negative societal impacts, the authors could also discuss possible mitigation strategies (e.g., gated release of models, providing defenses in addition to attacks, mechanisms for monitoring misuse, mechanisms to monitor how a system learns from feedback over time, improving the efficiency and accessibility of ML).
    \end{itemize}
    
\item {\bf Safeguards}
    \item[] Question: Does the paper describe safeguards that have been put in place for responsible release of data or models that have a high risk for misuse (e.g., pretrained language models, image generators, or scraped datasets)?
    \item[] Answer: \answerNA{} 
    \item[] Justification: The paper poses no such risks.
    \item[] Guidelines:
    \begin{itemize}
        \item The answer NA means that the paper poses no such risks.
        \item Released models that have a high risk for misuse or dual-use should be released with necessary safeguards to allow for controlled use of the model, for example by requiring that users adhere to usage guidelines or restrictions to access the model or implementing safety filters. 
        \item Datasets that have been scraped from the Internet could pose safety risks. The authors should describe how they avoided releasing unsafe images.
        \item We recognize that providing effective safeguards is challenging, and many papers do not require this, but we encourage authors to take this into account and make a best faith effort.
    \end{itemize}

\item {\bf Licenses for existing assets}
    \item[] Question: Are the creators or original owners of assets (e.g., code, data, models), used in the paper, properly credited and are the license and terms of use explicitly mentioned and properly respected?
    \item[] Answer: \answerYes{} 
    \item[] Justification: We have cited the original paper that produced the dataset.
    \item[] Guidelines:
    \begin{itemize}
        \item The answer NA means that the paper does not use existing assets.
        \item The authors should cite the original paper that produced the code package or dataset.
        \item The authors should state which version of the asset is used and, if possible, include a URL.
        \item The name of the license (e.g., CC-BY 4.0) should be included for each asset.
        \item For scraped data from a particular source (e.g., website), the copyright and terms of service of that source should be provided.
        \item If assets are released, the license, copyright information, and terms of use in the package should be provided. For popular datasets, \url{paperswithcode.com/datasets} has curated licenses for some datasets. Their licensing guide can help determine the license of a dataset.
        \item For existing datasets that are re-packaged, both the original license and the license of the derived asset (if it has changed) should be provided.
        \item If this information is not available online, the authors are encouraged to reach out to the asset's creators.
    \end{itemize}

\item {\bf New Assets}
    \item[] Question: Are new assets introduced in the paper well documented and is the documentation provided alongside the assets?
    \item[] Answer: \answerNA{} 
    \item[] Justification: The paper does not release new assets.
    \item[] Guidelines:
    \begin{itemize}
        \item The answer NA means that the paper does not release new assets.
        \item Researchers should communicate the details of the dataset/code/model as part of their submissions via structured templates. This includes details about training, license, limitations, etc. 
        \item The paper should discuss whether and how consent was obtained from people whose asset is used.
        \item At submission time, remember to anonymize your assets (if applicable). You can either create an anonymized URL or include an anonymized zip file.
    \end{itemize}

\item {\bf Crowdsourcing and Research with Human Subjects}
    \item[] Question: For crowdsourcing experiments and research with human subjects, does the paper include the full text of instructions given to participants and screenshots, if applicable, as well as details about compensation (if any)? 
    \item[] Answer: \answerNA{} 
    \item[] Justification: The paper does not involve crowdsourcing or research with human subjects.
    \item[] Guidelines:
    \begin{itemize}
        \item The answer NA means that the paper does not involve crowdsourcing nor research with human subjects.
        \item Including this information in the supplemental material is fine, but if the main contribution of the paper involves human subjects, then as much detail as possible should be included in the main paper. 
        \item According to the NeurIPS Code of Ethics, workers involved in data collection, curation, or other labor should be paid at least the minimum wage in the country of the data collector. 
    \end{itemize}

\item {\bf Institutional Review Board (IRB) Approvals or Equivalent for Research with Human Subjects}
    \item[] Question: Does the paper describe potential risks incurred by study participants, whether such risks were disclosed to the subjects, and whether Institutional Review Board (IRB) approvals (or an equivalent approval/review based on the requirements of your country or institution) were obtained?
    \item[] Answer: \answerNA{} 
    \item[] Justification: The paper does not involve crowdsourcing nor research with human subjects.
    \item[] Guidelines:
    \begin{itemize}
        \item The answer NA means that the paper does not involve crowdsourcing nor research with human subjects.
        \item Depending on the country in which research is conducted, IRB approval (or equivalent) may be required for any human subjects research. If you obtained IRB approval, you should clearly state this in the paper. 
        \item We recognize that the procedures for this may vary significantly between institutions and locations, and we expect authors to adhere to the NeurIPS Code of Ethics and the guidelines for their institution. 
        \item For initial submissions, do not include any information that would break anonymity (if applicable), such as the institution conducting the review.
    \end{itemize}

\end{enumerate}

\end{document}